\documentclass[3p,,preprint,review,12pt]{elsarticle}

\usepackage{tabulary,xcolor}
\usepackage{amsfonts,amsmath,amssymb}

\journal{Applied Soft Computing}

\usepackage[T1]{fontenc}
\makeatletter
\let\save@ps@pprintTitle\ps@pprintTitle
\def\ps@pprintTitle{\save@ps@pprintTitle\gdef\@oddfoot{\footnotesize\itshape \null\hfill\today}}
\def\hlinewd#1{%
  \noalign{\ifnum0=`}\fi\hrule \@height #1%
  \futurelet\reserved@a\@xhline}
\def\tbltoprule{\hlinewd{.8pt}\\[-12pt]}
\def\tblbottomrule{\noalign{\vspace*{6pt}}\hline\noalign{\vspace*{2pt}}}
\def\tblmidrule{\noalign{\vspace*{6pt}}\hline\noalign{\vspace*{2pt}}}

\AtBeginDocument{\ifNAT@numbers \biboptions{sort&compress}\fi}
\makeatother

\usepackage{url,multirow,morefloats,floatflt,cancel,tfrupee}
\makeatletter

\AtBeginDocument{\@ifpackageloaded{textcomp}{}{\usepackage{textcomp}}}
\makeatother
\usepackage{colortbl}
\usepackage{xcolor}
\usepackage{pifont}
\usepackage[nointegrals]{wasysym}
\makeatletter

\def\mcWidth#1{\csname TY@F#1\endcsname+\tabcolsep}

\def\cAlignHack{\rightskip\@flushglue\leftskip\@flushglue\parindent\z@\parfillskip\z@skip}
\def\rAlignHack{\rightskip\z@skip\leftskip\@flushglue \parindent\z@\parfillskip\z@skip}

\@ifundefined{etal}{}{}

\usepackage{ifxetex}
\ifxetex\else\if@twocolumn\@ifpackageloaded{stfloats}{}{\usepackage{dblfloatfix}}\fi\fi

\AtBeginDocument{
\expandafter\ifx\csname eqalign\endcsname\relax
\def\eqalign#1{\null\vcenter{\def\\{\cr}\openup\jot\m@th
  \ialign{\strut$\displaystyle{##}$\hfil&$\displaystyle{{}##}$\hfil
      \crcr#1\crcr}}\,}
\fi
}

\AtBeginDocument{%
  \@ifpackageloaded{endfloat}%
   {\renewcommand\efloat@iwrite[1]{\immediate\expandafter\protected@write\csname efloat@post#1\endcsname{}}}{\newif\ifefloat@tables}%
}%

\def\BreakURLText#1{\@tfor\brk@tempa:=#1\do{\brk@tempa\hskip0pt}}
\let\lt=<
\let\gt=>
\def\processVert{\ifmmode|\else\textbar\fi}

\@ifundefined{subparagraph}{
\def\subparagraph{\@startsection{paragraph}{5}{2\parindent}{0ex plus 0.1ex minus 0.1ex}%
{0ex}{\normalfont\small\itshape}}%
}{}

\newcommand\role[1]{\unskip}
\newcommand\aucollab[1]{\unskip}
  
\@ifundefined{tsGraphicsScaleX}{\gdef\tsGraphicsScaleX{1}}{}
\@ifundefined{tsGraphicsScaleY}{\gdef\tsGraphicsScaleY{.9}}{}
\def\checkGraphicsWidth{\ifdim\Gin@nat@width>\linewidth
	\tsGraphicsScaleX\linewidth\else\Gin@nat@width\fi}

\def\checkGraphicsHeight{\ifdim\Gin@nat@height>.9\textheight
	\tsGraphicsScaleY\textheight\else\Gin@nat@height\fi}

\def\fixFloatSize#1{}
\let\ts@includegraphics\includegraphics

\def\inlinegraphic[#1]#2{{\edef\@tempa{#1}\edef\baseline@shift{\ifx\@tempa\@empty0\else#1\fi}\edef\tempZ{\the\numexpr(\numexpr(\baseline@shift*\f@size/100))}\protect\raisebox{\tempZ pt}{\ts@includegraphics{#2}}}}

\AtBeginDocument{\def\includegraphics{\@ifnextchar[{\ts@includegraphics}{\ts@includegraphics[width=\checkGraphicsWidth,height=\checkGraphicsHeight,keepaspectratio]}}}

\DeclareMathAlphabet{\mathpzc}{OT1}{pzc}{m}{it}

\def\URL#1#2{\@ifundefined{href}{#2}{\href{#1}{#2}}}

\def\UrlOrds{\do\*\do\-\do\~\do\'\do\"\do\-}%
\g@addto@macro{\UrlBreaks}{\UrlOrds}

\edef\fntEncoding{\f@encoding}

\makeatother

\newif\ifmultipleabstract\multipleabstractfalse%
\usepackage{amssymb}

\usepackage{lineno}

\usepackage{float}

\begin{document}

\begin{frontmatter}
	
\title{
    \textbf{Unsupervised Textile Defect Detection using Convolutional Neural Networks}    
}
    
\author[]{Imane Koulali\corref{c-eb13a6755e4e}}
\ead{imane.koulali@isikun.edu.tr}\cortext[c-eb13a6755e4e]{Corresponding author.}
\author[]{M. Taner Eskil}
\ead{taner.eskil@isikun.edu.tr}
    
\address{Department of Computer Science Engineering\unskip, 
    I{\c{s}}{\i}k University\unskip, B{\"{u}}y{\"{u}}kdere Ave.  No: 106, Maslak\unskip, 34398\unskip, Istanbul\unskip, Turkey}

\begin{abstract}
In this study, we propose a novel motif-based approach for unsupervised textile anomaly detection that combines the benefits of traditional convolutional neural networks with those of an unsupervised learning paradigm. It consists of five main steps:  preprocessing, automatic pattern period extraction, patch extraction, features selection and anomaly detection. This proposed approach uses a new dynamic and heuristic method for feature selection which avoids the drawbacks of initialization of the number of filters (neurons) and their weights, and those of the backpropagation mechanism such as the vanishing gradients, which are common practice in the state-of-the-art methods. The design and training of the network are performed in a dynamic and input domain-based manner and, thus, no ad-hoc configurations are required. Before building the model, only the number of layers and the stride are defined. We do not initialize the weights randomly nor do we define the filter size or number of filters as conventionally done in CNN-based approaches. This reduces effort and time spent on hyper-parameter initialization and fine-tuning. Only one defect-free sample is required for training and no further labeled data is needed. The trained network is then used to detect anomalies on defective fabric samples. We demonstrate the effectiveness of our approach on the Patterned Fabrics benchmark dataset. Our algorithm yields reliable and competitive results (on recall, precision, accuracy and f1-measure) compared to state-of-the-art unsupervised approaches, in less time, with efficient training in a single epoch and a lower computational cost.
\end{abstract}


\end{frontmatter}
    
\section{Introduction}
In the modern textile industry, defect detection is a critical stage in effective fabric quality control. A roll of textile is priced with respect to several variables; these include the defects' size, their number and their locations on the roll. 

Defect detection has historically been carried out by human experts solely relying on visual inspection. This type of recognition is not only time-consuming but also suffers from the high intensity of labor and human oversight. Fatigue and inattentiveness indeed hamper effective quality control, especially when defects are small and have a low contrast. Moreover, these results are mostly subjective and unquantifiable. The effectiveness of this manual approach is hence low which results in a high defect rate. Yielded results are consequently unreliable, mainly due to the versatility of the human factor. Ergo, computerized and automated inspection techniques to detect defects are vital. 

Over the years, many methods have been put forward to address this problem. In the literature, these approaches broadly fall under two main categories; motif-based and non-motif based approaches. In the non-motif based group, approaches can further be classified into six categories; statistical, spectral, model-based, learning-based, structural and hybrid methods. 

Statistical methods aim to detect defective regions based on statistical properties such as regularity, uniformity and similarity. This category of methods includes the co-occurence
matrix\unskip~\cite{981091:21201186, 981091:21201184}, the auto-correlation metric\unskip~\cite{981091:21238444}, mathematical morphology\unskip~\cite{981091:21201195} and fractal dimension methods \unskip~\cite{981091:21201182}. These methods are usually suitable for homogeneous textures.

Spectral-based methods are another set of approaches which detect defects in the frequency domain. Typical spectral-based approaches are Fourier transform\unskip~\cite{981091:21201190}, Gabor transformation\unskip~\cite{981091:21201199} and Wavelet-based methods \unskip~\cite{981091:21201168}. Albeit their effectiveness, these approaches come with a considerably high computational cost.

Model-based approaches represent the normal texture with probability distribution models such as Gaussian mixture models \unskip~\cite{981091:21201219}. They then consider nonconforming areas as defects by utilizing the constructed model. Together with structural methods, these approaches are highly data dependent and are therefore less common.

Machine learning based approaches proved to be promising for defect detection. In these methods, features are extracted from training images and later used to distinguish damaged areas from the defect free ones on the sample being tested. Nonetheless, most of the methods that perform well in fabric defect detection tasks are supervised learning approaches, they require manually labeled samples for model training which is costly and challenging to obtain.

In the recent years, use of Deep Learning (DL) algorithms rapidly proliferated in almost every domain, including anomaly detection in general and textile defect detection in particular\unskip~\cite{981091:21201220},\unskip~\cite{981091:21201222},\unskip~\cite{981091:21201221}. These DL based algorithms represent data with numerous levels of abstraction by use of many processing layers. The image is passed through the pipeline of the trained model and the segmentation is contingent on the deep features produced. These features often outperform handmade and pre-defined feature groups. DL approaches not only gained popularity but also became the state-of-the-art in the computer vision field due to their feature learning capability. Among all deep learning techniques, Convolutional Neural Networks (CNNs) have shown to be better in a plentiful of problems, including visual object recognition and detection. 

Despite the variety of neural network based architectures proposed in the literature, common decisive steps in the great majority of CNN-based approaches are to design the network itself and find a suitable training strategy for the designed architecture. The first stage involves pre-configuring several hyperparameters for each layer to obtain a satisfying network architecture, namely the number of filters along with their size and how they will be initialized, the activation function, the learning rate and the stride. This pre-configuration is solely guided by empirical trial and error which is ad-hoc and time-costly. Spanning the entire search space requires numerous neurons in every layer. This increases the computational cost of the CNN and is prone to overfitting.

Most state-of-the-art CNN networks\unskip~\cite{981091:21201220},\unskip~\cite{981091:21201221},\unskip~\cite{981091:21201181},\unskip~\cite{981091:21201173}, \unskip~\cite{981091:21201172} rely on backpropagation to train the network, as initially proposed by S. Linnainmaa\unskip~\cite{981091:21231864}, by incrementally updating weights with respect to the partial derivative of the error. The gradient-based nature of this algorithm, alone, comes at a price. One of its drawbacks, is that converging to meaningful features requires multiple epochs. Indeed, should we have taken large steps, we would have faced the risk of divergence. In addition, backpropagation is prone to the issue of vanishing gradients. Another limitation is the fact that shallow layers' training becomes difficult since the gradient's value decreases in the direction of the first layer. Since correctly distributing and backpropagating the error to shallow layers is challenging, the quality of the features obtained in these layers is hindered, which negatively impacts the task of feature detection.  

Most works in fabric defect detection chose to utilize non-motif based approaches. Motif-based methods, nonetheless, generally have better generality. A fabric is composed of a basic fundamental unit, called a motif\unskip~\cite{981091:21231851}, that is reproduced on the fabric by rules of symmetry. These elementary motifs are considered as manipulation units by motif-based approaches.

To overcome the dependency on the presence of reliable and sufficient training data, and by capitalizing on motif repetitiveness in fabric, we propose a competitive unsupervised motif-based approach for fabric defect detection. This approach integrates the strengths of the Convolutional Neural Network architecture with the advantages of an unsupervised learning paradigm. It also isn't subject to the limitations that are posed by backpropagation.

In our approach, both the design of the deep convolutional network architecture and its training are performed based solely on observations in the input domain, without any ad-hoc configurations. Before the model is built, we only define the number of layers and the stride. We don't define the filter size, number of filters and neither do we initialize the weights randomly as conventionally done in CNNs, which reduces the time and effort spent on hyper-parameter initialization. For every convolutional layer, neurons are created and initialized on a need basis as features are gradually discovered from the input domain, and until the sufficient number of features is reached. We also do not require any labeled defective samples for training. Therefore, our approach enables fast and efficient training with low computational cost and high detectability. It yields reliable results and has a high accuracy rate.

The remaining sections are organized as follows. Section \textbf{2} describes the related work, with a focus on unsupervised methods. In section \textbf{3}, the proposed approach is described in detail. The experimental setup and implementation details are provided in section \textbf{4}. In section \textbf{5}, we demonstrated the experimental results of our approach. Finally, in section \textbf{6} we conclude by summarizing the content of this paper and discussing future work.

\section{Related work}

With the recent years’ success of CNNs, these networks have been widely used in the computer vision field in general, and in object detection and image segmentation tasks, in particular. These approaches are particularly suitable for these applications and have yielded revolutional results due to their ability to automatically extract useful features from data without the need for hand-crafted features. CNN-based methods have been used both for fabric defect classification and detection as well. 

Several successful network architectures in other detection and classification tasks have been adapted and utilized for fabric defects’ detection and classification.  
For instance, J.Jing et al. (2017) \unskip~\cite{Jing2017YarndyedFD} proposed a modified AlexNet for yarn-dyed fabric defect detection, H. Zhang et al. (2018) \unskip~\cite{Zhang2018YarndyedFD} introduced a defect location algorithm based on YOLO and Liu et al. (2019) \unskip~\cite{Liu2019FabricDD} implemented a computational efficient CNN-based framework (named as YOLO-LFD) that uses K-means reduction in feature dimensions and provided a compression of YOLO-v2 \unskip~\cite{Redmon2017YOLO9000BF}.

Wei et al. (2019) \unskip~\cite{Wei2019ANM} proposed a hybrid approach that uses CNN with compressed sensing  for defect classification, based on small sample sizes. The reported accuracy of the implemented approach is 97.9\%.

Wu et al. (2020) \unskip~\cite{Wu2020AutomaticFD} suggested an automated composite interpolating feature pyramid (CI-FPN) approach, with deformable convolution filters, for fabric defect detection. This method uses a cascaded guided region proposal network (CG-RPN) to sort the detected defective regions. The approach was tested on the FBDF dataset and its accuracy reached 80.4\% accuracy.

Jeyaraj et al. (2020) \unskip~\cite{Jeyaraj2020EffectiveTQ} presented a ResNet512-based CNN for textile defect detection. The classification accuracy and precision reported on the TILDA dataset are, respectively,  96.5\% and 98.5\%. 
Wen et al. (2020) \unskip~\cite{Wen2020CNNbasedMF} implemented a PETM-CNN algorithm which consists of a Patch Extractor algorithm (PE) and a Triplet Metric CNN (TM-CNN) module that predicts patches’ labels. Their method was mainly designed for the detection of minor defects and achieved 79.32\% accuracy.

Jin et al. (2020) \unskip~\cite{981091:21201220} proposed Mobile-Unet, a CNN-based method for fabric defect detection, where they replaced the encoding part of U-net \unskip~\cite{Falk2018UNetDL} with MobileNetV2 \unskip~\cite{Sandler2018MobileNetV2IR} Their implemented method is efficient as it achieves 99.75\% and 98.80\% accuracy on the YID and FID benchmark datasets, respectively. The approach is compared with FCN\unskip~\cite{981091:21201181}, SegNet\unskip~\cite{981091:21201173}, U-Net\unskip~\cite{981091:21201172}, PTIP\unskip~\cite{981091:21201221} on the same fabric datasets and achieves the best performance both in terms of accuracy and of intersection over union metric.

Due to the need for automatic methods for fabric defect detection, in order to better tackle the needs of the textile industry, many unsupervised approaches have been proposed. Since our method is an unsupervised method, we mainly focus on comparing our results with previous unsupervised approaches. As expected, supervised approaches naturally achieve better performances than the unsupervised methods, but at the expense of computational and time complexity, especially in the training phase. Although unsupervised approaches may achieve a comparable recall score with respect to supervised approaches, their precision and f1-score tend to be lower. 
In the following, we outline some of the state-of-the-art unsupervised approaches developed for fabric defect detection over the years. We later compare their performance against that of our approach in the Results section \textbf{5}.

H. Ngan et al. (2005)\unskip~\cite{981091:21201168}  developed Wavelet preprocessed golden image subtraction (WGIS), a Wavelet-transform based method for fabric defect detection. A first-level approximation is extracted with Haar WT, then GIS is applied on the approximated subimage to detect defects. 

H. Ngan and G. Pang (2006)\unskip~\cite{981091:21201212} introduced a new application of Bollinger Bands (BB) for patterned fabric defect detection, with the concept of the moving average. This method is efficient, it nonetheless fails to detect small defects and defects with only light color differences from the background.

H. Ngan and G. Pang (2007)\unskip~\cite{981091:21201213} proposed a fabric defect-detection method based on regularity. This method builds on enhancing and segmenting defective regions through moving averages and standard deviations. The method's strong suit is its simplicity. However, it is not as successful in detecting defects near the borders.

V. Asha et al. (2011)\unskip~\cite{981091:21201199} utilized texture periodicity and Gabor kernels to analyze faulty blocks by utilizing Ward's clustering. The most important limitation of this technique is that it cannot be used to validate defect-free images.

Anitha et al. (2013)\unskip~\cite{981091:21201202} proposed a method that can succeed in characteristic extraction, by analyzing cloth defect images using Gabor wavelet network and an independent component. This approach is only valid for patterned fabric containing a parallelogram, rectangle, square or a hexagonal shape. These types of fabric patterns are the most common in the textile domain, they are hence easy to learn. On the contrary, other less common and irregular shapes along with non-uniform brightness and low contrast raise an unanswered challenge.

M. Ng et al. (2014) \unskip~\cite{981091:21201214} suggested an approach based on image decomposition (ID). The model decomposes the image into a cartoon structure that contains defective areas of the input image and a texture structure. 

C. Tsang et al. (2016) \unskip~\cite{981091:21201215} introduced the Elo rating method (ER). This approach uses an Elo point matrix to match between partitions of standard size from the input image. Strong defects result in high Elo points whereas light defects result in low Elo points. This method has a drawback in Netting Multiple and Hole defects.

G. Hu et al. (2015)\unskip~\cite{981091:21201189} proposed an unsupervised approach for defect detection using Fourier analysis and Wavelet shrinkage. The gradient distribution of the input image's Fourier spectrum is analyzed and components representing defect-free regions are eliminated using Discrete Fourier Transform (DFT) and zero-masking. As a result, the Inverse Fourier Transform (IDFT) contains almost only defect-free information. The Fourier restored residual image is then denoised using Discrete Wavelet Transform (DWT). Reconstructions from the approximation and wavelet coefficients are thresholded which results in two binary filtered images, containing respectively low-frequency information and high-frequency information of the defect. Fusion is then performed to merge these two images and obtain a final segmentation of the defect for the input image. The advantage of this method is that it requires no reference images and performs well on plain textile. However, changes in illumination, scale or orientation of the input images result in poor results. 

J. Zhou et al. (2016)\unskip~\cite{981091:21201191} introduced a fully unsupervised fabric defect detection method by using local patch approximation. The first step of their approach is the extraction of overlapping patches from the input image without distinguishing between defect-free and defective patches. After that, a threshold is set on the Euclidian distance between every patch and the data center which allows to cast out outlier patches, while only keeping defect-free patches in a dictionary. Following that, an abnormality map is constructed by subtracting patches previously stored from the input image, then segmentation is performed on the output in order to obtain a binary image of defective areas. Similarly to the previous approach, this method does not require reference images. However, due to this, the recall rate is sacrificed in favor of the detection rate which leads to a high false detection rate.

A. A. Hamdi et al. (2018)\unskip~\cite{981091:21201201} proposed an automatic unsupervised patterned fabric defect detection algorithm that uses standard division filtering and k-means clustering. In this approach, the filtering operation highlights the defected area using a small window while the clustering is used to classify the different defected blocks in fabric images. While this method achieves good accuracy, many preprocessing steps are necessary to obtain good results. Indeed, changes in contrast, histogram equalization along with two different types of filtering are applied to images to ensure a good performance. 

More recently, there has been a growing interest in unsupervised deep learning and neural networks-based approaches which combine the advantages that deep learning offers along with the benefits of the unsupervised paradigm. 

Y. Li et al. (2016)\unskip~\cite{981091:21201196} proposed a defect detection system using Fisher Criterion-based Stacked Denoising Auto-Encoders (FCSDA) to classify fabric patches into defect-free and defective regions. They gathered a dataset of both categories and divided the images into patches of the same size while manually labeling them in order to train the FCSDA, which consists of four hidden layers and a softmax output layer. A set of patches is classified through the FCSDA and the residual between the reconstructed image and the defective patch is calculated. Then, the defect is located through thresholding. The residual error for the output layer is computed and it is iterated for other layers using the fisher criterion as a loss function for the SDA. This method can reach a good detection accuracy, especially on plain fabric. Nonetheless, the pretraining subprocess is very time-consuming due to the complexity of the FCSDA framework. Also, this method requires that each Fabric be photographed from different angles. Additionally, several hyper-parameters should be initialized such as the learning rate and the number of neurons per layer.

Seker et al. (2017)\unskip~\cite{981091:21201185} implemented a deep learning method for fabric defect detection using autoencoders. They focused on tuning the hyperparameters of the autoencoder-based model in order to improve feature extraction and get better classification results. Their method showcased a high detection accuracy. However, their dataset was created manually under ideal and perfect shooting conditions. Furthermore, this method requires tuning many hyperparameters and only works on woven fabrics.

S. Mei et al. (2018)\unskip~\cite{981091:21201198} proposed an unsupervised fabric defect detection model based on Multi-Scale Convolutional Denoising Auto-Encoder networks (MSCDAE). Patches of different sizes are randomly extracted and every pixel is later characterized based on its local neighborhood. The model is trained on defect-free samples. Hence, the filters in the trained model react differently to defective patches than to defect-free patches, large response vs. low, respectively. Thus, by measuring the residual between the response and the original image, defective areas can be detected after gathering results from the top of the pyramid layers. This method achieves good results on simple fabrics but fails to correctly detect defects on complex textures. Additionally, several preprocessing steps are required such as illumination normalization, and noise corruption of the patches. 

L. Liu et al. (2018)\unskip~\cite{981091:21201197} introduced a novel fabric defect classification method using Extreme Learning Machines (ELM)~based on a single hidden layer feedforward network algorithm. In their approach, defect segmentation was divided into four steps, namely feature extraction, ELM classifier training, Bayesian probability fusion and defect segmentation. Albeit the great performance of this method on common woven fabrics, it can't yield good results on other fabric types and textures. 

G. Hu et al. (2019)\unskip~\cite{981091:21201204} presented a method for detecting the defects using a Deep Convolutional Generative
Adversarial Network (DCGAN) and ConvNet models. This approach first consists in creating a query from the defected image's space and subtracting it from the original image. Then, defect occurrence probabilities are calculated. This is followed by several steps to increase the accuracy rate. Although this method follows an unsupervised scheme, it needs defect free fabric images in order to train the network model. Moreover, both generating a reconstructed image and calculating defect probabilities require significant computational power. 

H. Tian and F. Li (2019)\unskip~\cite{981091:21201169} proposed an iterative autoencoder-based approach with cross-patch similarity. For every testing patch, this method finds the similar non-defective patches and weighs their corresponding latent variables to modify the original latent variable of the patch. It then detects defected areas from the reconstruction residual with a decoder. This method, however, requires several iterations to converge.

The main contributions of our work are:
  
  \begin{enumerate}
  \item \relax We only need one defect-free sample per fabric type in our training phase, the difficulties of finding sufficient defective samples are thus avoided without performing any data augmentation;
  \item \relax We don't necessitate pictures of different angles in our defect detection scheme;
  \item \relax We avoided applying any preprocessing steps to our images, except for histogram equalization;
  \item \relax Our unsupervised training does not have large computational or time requirements and no pretraining is needed;
  \item \relax We reduced the number of hyperparameters in our approach compared to classical CNN methods, in order to avoid hyper-parameter tuning which is time-consuming; 
  \item \relax Our training is performed without the backpropagation mechanism which allows us to converge to meaningful features in a single epoch and in less time than other deep learning based approaches;
  \item \relax Only one hidden layer is usually enough to achieve comparable results and no random initialization of weights is done;
  \item \relax Our method achieves an accuracy that is as high as the state-of-the-art methods with higher recall, precision and f1-score.
  \end{enumerate}

\section{Methodology}

\bgroup
\fixFloatSize{Figures/Figure1.jpg}
\begin{figure*}[!htbp]
\centering \makeatletter\IfFileExists{Figures/Figure1.jpg}{\includegraphics{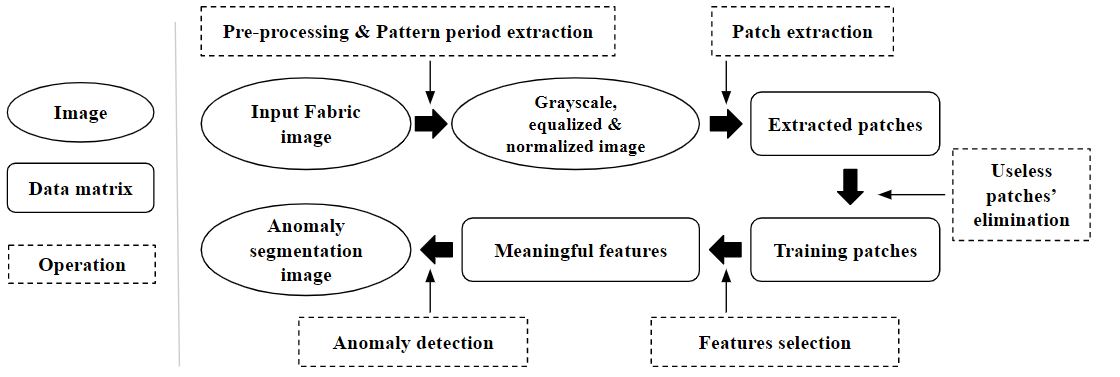}}{}
\makeatother 
\caption{{Schematic description of anomaly detection stages}}
\label{f-ff9ab3b9d830}
\end{figure*}
\egroup

Our approach consists of five main steps, which can be organized in three main phases : 
\begin{itemize}
\item The first phase corresponds to steps carried out before training, namely preprocessing (section \textbf{3.1.}), automatic pattern period extraction (section \textbf{3.2.}) and patch extraction (section \textbf{3.3.}). 
\item The second phase corresponds to training. After patch extraction, training is done in an unsupervised manner and mainly consists of the features' selection step (section \textbf{3.4.}). 
\item The last phase, which is testing, corresponds to the anomaly detection step (section \textbf{3.5.}).
\end{itemize}
Figure~\ref{f-ff9ab3b9d830} shows the overall detection scheme of our approach. 

\subsection{Preprocessing}
Fabric images' acquisition is usually performed with digital cameras that may suffer from poor contrast, which impacts the detection. To mitigate these undesirable effects, the contrast of the input images needs to be adjusted.  

We have tested several contrast enhancement approaches like gamma correction, linear adjustment, adaptive histogram equalization and histogram equalization. The simple histogram equalization yielded the best results in terms of final scores of our detection approach. It is hence used for both training and testing images.

Once histogram equalization is completed, we divide the pixel values by 255 in order to normalize them between [0,1], followed by data standardization as shown in Equation~(\ref{dfg-a7c71ef1aa50}) and Equation~(\ref{dfg-b1debec2d148}). 
\let\saveeqnno\theequation
\let\savefrac\frac
\def\dispfrac{\displaystyle\savefrac}
\begin{eqnarray}
\let\frac\dispfrac
\gdef\theequation{1}
\let\theHequation\theequation
\label{dfg-a7c71ef1aa50}
\begin{array}{@{}l}I\;\leftarrow I\;-\;\mu\end{array}
\end{eqnarray}
\global\let\theequation\saveeqnno
\addtocounter{equation}{-1}\ignorespaces 

\let\saveeqnno\theequation
\let\savefrac\frac
\def\dispfrac{\displaystyle\savefrac}
\begin{eqnarray}
\let\frac\dispfrac
\gdef\theequation{2}
\let\theHequation\theequation
\label{dfg-b1debec2d148}
\begin{array}{@{}l}I\;\leftarrow\frac I\sigma \;\end{array}
\end{eqnarray}
\global\let\theequation\saveeqnno
\addtocounter{equation}{-1}\ignorespaces 
 With \textbf{\textit{\ensuremath{\mu }}}and \textbf{\textit{\ensuremath{\sigma  }}} being the mean and the standard deviation of \textbf{\textit{I}}, respectively.

\subsection{The period of the pattern}In order to detect defects correctly, the determining the patch size of the patches is critical. Extracting the period of the pattern from the input image has proven to be powerful for the analysis of defective images. Defects can be detected in an easier way if the input image is divided into patches that have the same size following the periodicity. 

In order to obtain the period for the input image, we calculate the row-wise average and the column-wise average of the intensity values of the image. Let \textbf{\textit{I}} be a fabric image of size \textbf{\textit{W x H}}, the averages are then obtained byEquation~(\ref{dfg-9c08b1b9cf3d}) and Equation~(\ref{dfg-63289e1c51ca}) below.

\let\saveeqnno\theequation
\let\savefrac\frac
\def\dispfrac{\displaystyle\savefrac}
\begin{eqnarray}
\let\frac\dispfrac
\gdef\theequation{3}
\let\theHequation\theequation
\label{dfg-9c08b1b9cf3d}
\begin{array}{@{}l}R(k)\;=\;\frac1W\sum_{\;j=1}^{W}\;I(k,j)\;;\;k\;=1:W\end{array}
\end{eqnarray}
\global\let\theequation\saveeqnno
\addtocounter{equation}{-1}\ignorespaces 
\let\saveeqnno\theequation
\let\savefrac\frac
\def\dispfrac{\displaystyle\savefrac}
\begin{eqnarray}
\let\frac\dispfrac
\gdef\theequation{4}
\let\theHequation\theequation
\label{dfg-63289e1c51ca}
\begin{array}{@{}l}C(k)\;=\;\frac1H\;\sum_{i=1}^{H}\;I(k,i)\;;\;k\;=1:H\end{array}
\end{eqnarray}
\global\let\theequation\saveeqnno
\addtocounter{equation}{-1}\ignorespaces 
Using auto-correlation on each of the two obtained vectors allows us to extract the period along each of these directions by measuring the distances between the successive peaks.  Finally, taking the median of these distances has proven to be a good way to get the final periodicity value. These steps are illustrated in Figure~\ref{f-a502e190eb4d}.

\bgroup
\fixFloatSize{Figures/Figure2.jpg}
\begin{figure*}[!htbp]
\centering \makeatletter\IfFileExists{Figures/Figure2.jpg}{\includegraphics{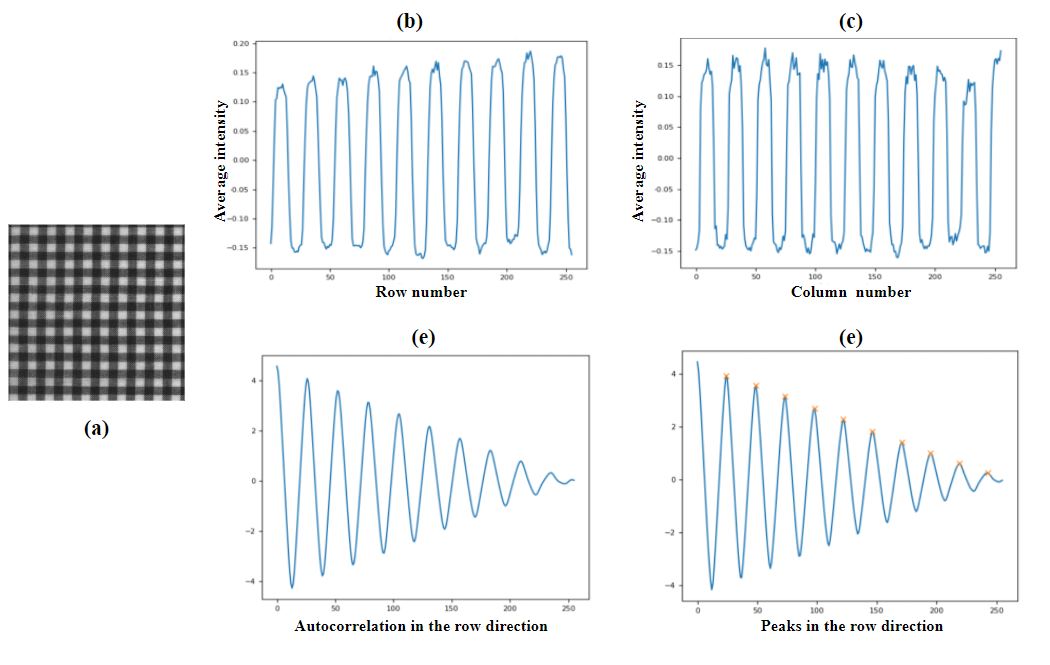}}{}
\makeatother 
\caption{{\textbf{(a)} Box-patterned defect-free image; \textbf{(b)} Row-wise average vector; \textbf{(c)} Column-wise average vector; \textbf{(d)} Auto-correlation for the row direction; \textbf{(e)} Successive peaks detection }}
\label{f-a502e190eb4d}
\end{figure*}
\egroup

\subsection{Extraction of the patches}
Given a \textbf{\textit{W x H}} input image \textit{\textbf{I}, \textbf{p x p}} patches are cropped from the image \textit{\textbf{I}} \textbf{\space }where the patch size \textit{\textbf{p}} corresponds to the period found in the previous section. This patch size will be referred to as the \textit{filter window size} or the \textit{filter\_size}, it corresponds to the receptive field of the neural network.

\bgroup
\fixFloatSize{Figures/Figure3.jpg}
\begin{figure*}[!htbp]
\centering \makeatletter\IfFileExists{Figures/Figure3.jpg}{\includegraphics{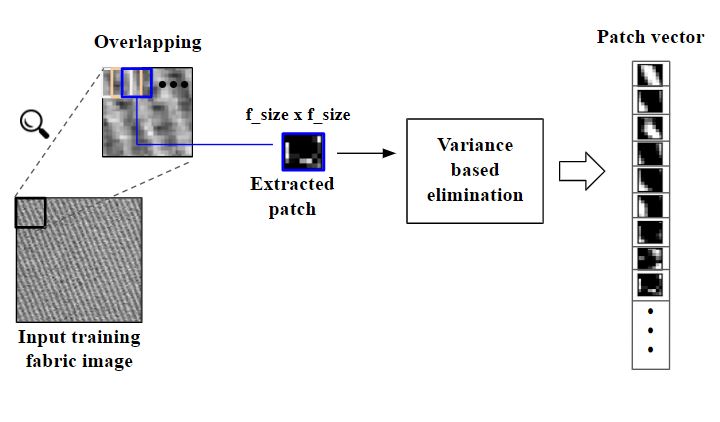}}{}
\makeatother 
\caption{{Illustration of the patch extraction step with overlapping division}}
\label{f-9b03f817d68d}
\end{figure*}
\egroup
As shown in Figure~\ref{f-9b03f817d68d}, these patches are extracted and stored in a row by row manner, using overlapping division. Patches thus share partial regions with their neighbors in column and rows directions. Let \textbf{\textit{o}} be the stride parameter that controls the overlap, the total number of patches in image \textbf{I }can be calculated using Equation~(\ref{dfg-9e4b67c89eba}) :
\let\saveeqnno\theequation
\let\savefrac\frac
\def\dispfrac{\displaystyle\savefrac}
\begin{eqnarray}
\let\frac\dispfrac
\gdef\theequation{3}
\let\theHequation\theequation
\label{dfg-9e4b67c89eba}
\begin{array}{@{}l}N_{patches\;}=\;\left[\frac{W-p}o\right].\left[\frac{H\;-p}o\right]\end{array}
\end{eqnarray}
\global\let\theequation\saveeqnno
\addtocounter{equation}{-1}\ignorespaces 
In the end, an array of shape \textit{(N\ensuremath{_{patches}}, p x p)} is obtained for every input image. In training, each of the patches extracted from the training image(s) is a candidate for the feature selection step that is detailed in section \textbf{3.4}. 

In the testing phase, patches are extracted the same way and are used for image segmentation (anomalous vs. defect-free). This stage will be explained in section \textbf{3.5}.

Classifying a pixel by taking local neighborhood information into consideration has proven to be more robust than using a single pixel at a time. Nonetheless, the size of this neighborhood is data dependent. In our approach, the neighborhood size which corresponds to the filter window size gets larger as we progress towards deeper layers, which allows us to capture features at different scales, and hereby detect defects of different scales later on, if need be. Layers can be added on a need basis, depending on the input domain. 

For deeper layers, patches are extracted from the feature maps obtained by correlating the features of the previous layer with the training reference image. This is illustrated in lines 4 to 6 of the algorithm in \textbf{Figure 4}, and detailed in section {3.4.}.

In some cases, some of these extracted patches do not contain any meaningful information. Therefore, before training and features selection, we start by performing a variance-based elimination in order to trim useless patches. If the variance of the pixels in a patch is inferior to a defined contrast threshold, the patch is discarded. Otherwise, it is kept as a candidate for the next step which is filters selection. 

Finally, these patches are shuffled to avoid feature selection being biased by the patches' position in the input image and to ensure that every data patch creates an independent impact on the weights. 

After this step, we obtain a matrix \textbf{\textit{P}} of shape \textit{(N\ensuremath{_{useful}}\ensuremath{_{patches}}, p x p)} containing all of the useful extracted patches that features will be selected from.

\bgroup
\fixFloatSize{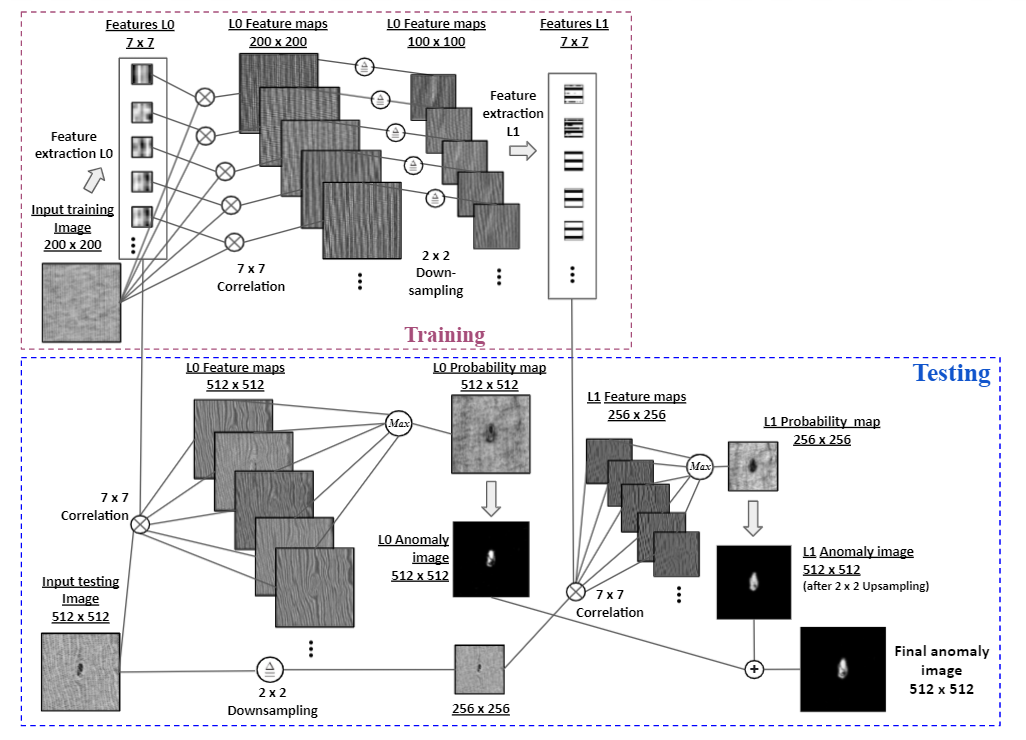}
\begin{figure}[!htbp]
\centering \makeatletter\IfFileExists{Figures/Figure4.PNG}{\includegraphics[width=\linewidth]{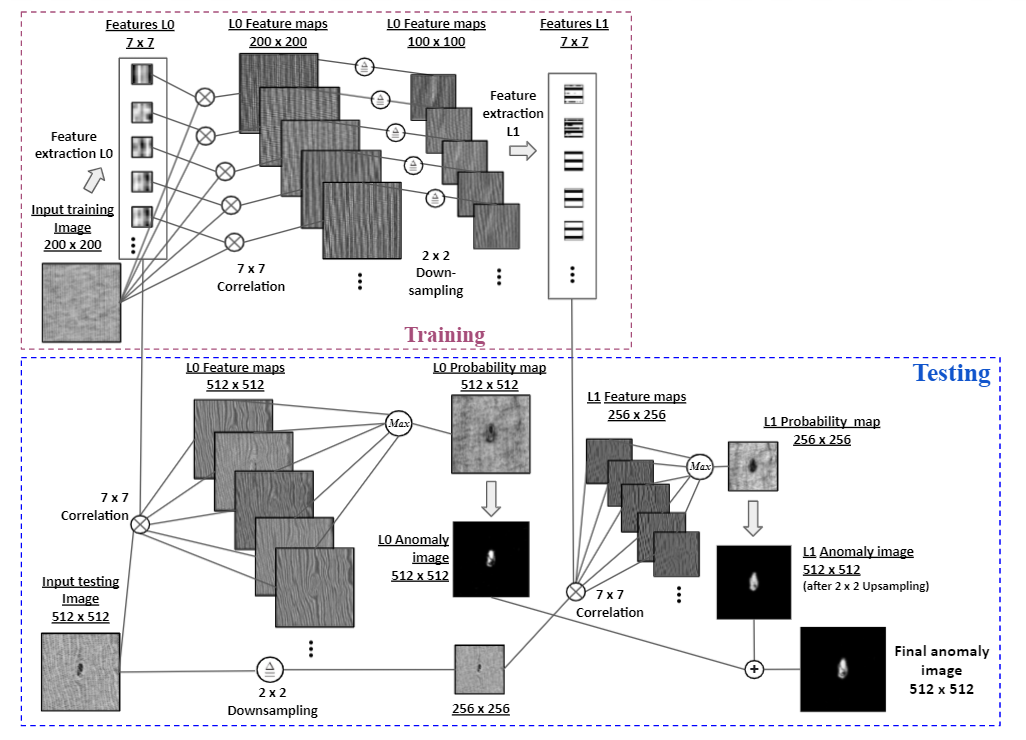}}{}
\makeatother 
\caption{{Illustration of the architecture of our network. An example with two layers.
Note : More layers can similarly be added on a need basis - depending on the input domain - and connected in the same way.}}
\label{f-TypicalNetworkArch}
\end{figure}
\egroup

\subsection{Selection of the features}
The training of our network mainly consists of this feature selection step. After this step is complete, the filter weights are deemed non-trainable and are used in the detection of anomalies step detailed in section \textbf{3.5} and no additional training is performed. Both the training and testing phases are shown in Figure~\ref{f-TypicalNetworkArch}

We do not provide the number of filters at any point, neither in model building nor in training. Features are dynamically selected progressively based on similarity calculation, depending on the input domain, following the filter discovery scheme performed by T. Erkoç and T. Eskil, An unsupervised approach for filter discovery in convolutional neural networks (under review).

As a consequence, the only parameter that needs to be set is a similarity threshold. The selection of features starts by calling the function \textit{TrainLayer} shown in Figure~\ref{f-6b30a3d493fd}. This function takes the current Layer \textbf{\textit{L}} and an array \textbf{\textit{I}} containing training images as inputs.

Layers are initially empty. The very first patch \textbf{\textit{P}}\textit{\textbf{\ensuremath{_{0}}}}\ensuremath{_{}} for every layer is always stored as the first feature \textbf{\textit{F}}\textit{\textbf{\ensuremath{_{0}}}}, since no filters exist at this point. Supporters \textbf{\textit{C}}\textit{\textbf{\ensuremath{_{0}}}} are the number of samples that contribute to the extracted feature, and they count as 1 for this newly added feature \textbf{\textit{F}}\textit{\textbf{\ensuremath{_{0}}}}.

For every other candidate patch \textbf{\textit{P}}\textit{\textbf{\ensuremath{_{i}}}}\ensuremath{_{}}, we calculate the similarity between this patch and all previously saved feature(s) by using dot product(s). These values are stored in a similarity array \textbf{\textit{S}}. What interests us the most is the feature \textbf{\textit{F}}\textit{\textbf{\ensuremath{_{j}}}} that is most similar to our candidate patch, so we find the maximum value of the calculated similarity scores, \textbf{\textit{S}}\textit{\textbf{\ensuremath{_{j}}}}. These similarity scores are in the [0,1] interval, and so is our similarity threshold. 

If \textbf{\textit{S}}\textit{\textbf{\ensuremath{_{j}}}} is lower than a similarity threshold, it means that the candidate patch is not similar enough to any of the already saved features and it needs to be saved as a new feature of the current layer since it contains new information that was not encountered so far. A new filter is thus created, and its supporter count is set to \textbf{\textit{C}}\textit{\textbf{\ensuremath{_{j}}}} \textbf{= 1}.

On the other hand, if \textbf{\textit{S}}\textit{\textbf{\ensuremath{_{j}}}} is higher than the similarity threshold, it means that we already saved a similar feature \textbf{\textit{F}}\textit{\textbf{\ensuremath{_{j}}}}. In this case, \textbf{\textit{P}}\textit{\textbf{\ensuremath{_{i}}}} becomes a supporter of \textbf{\textit{F}}\textit{\textbf{\ensuremath{_{j}}}}. Its supporters count \textbf{\textit{C}}\textit{\textbf{\ensuremath{_{j}}}} is thus incremented and the weights are updated accordingly.
\let\saveeqnno\theequation
\let\savefrac\frac
\def\dispfrac{\displaystyle\savefrac}
\begin{eqnarray}
\let\frac\dispfrac
\gdef\theequation{6}
\let\theHequation\theequation
\label{dfg-cb6d8d42480c}
\begin{array}{@{}l}F_j\;=\;F_j\;+\;\Delta\end{array}
\end{eqnarray}
\global\let\theequation\saveeqnno
\addtocounter{equation}{-1}\ignorespaces 

\let\saveeqnno\theequation
\let\savefrac\frac
\def\dispfrac{\displaystyle\savefrac}
\begin{eqnarray}
\let\frac\dispfrac
\gdef\theequation{7}
\let\theHequation\theequation
\label{dfg-f81ca8d8cde3}
\begin{array}{@{}l}C_j\;=\;C_j\;+1\;\end{array}
\end{eqnarray}
\global\let\theequation\saveeqnno
\addtocounter{equation}{-1}\ignorespaces 
with : $\Delta\;=\;(P_i\;-\;F_j)\;/\;(C_j\;+\;1)\; $

When all patches are processed, feature selection for the input layer is completed. For the input layer, the input image's patches are rawly fed to the training step. However, in order to add other deeper layers, extracted features of the previous layer are correlated with the previous input image and this correlation result is stored in \textbf{\textit{I}}.

The correlation performed is a cross-correlation with symmetrical boundary conditions that outputs a 2D array of the same size as the input image. Following that, \textit{stride x stride} downsampling is performed on this image which serves as the new \textbf{\textit{I}} input for the \textit{TrainLayer }function. 

Downsampling is performed by slicing the feature maps with respect to the \textit{stride}. It corresponds to taking elements with a step of \textit{stride} every time, across the height and width dimensions, from the array containing the feature maps. For instance, if the \textit{stride} parameter is set to 2, this would correspond to taking every other pixel from the image, both in the column and row directions. Every time, we would take one pixel, skip the other, and then take the next pixel, etc. In the end, we would have an image that has half the width and half the height of the original image.
As a result, performing patch extraction on the downsampled feature maps using the same initial  \textit{filter\_size},   would be as if we were doing patch extraction with an effective window twice as wide on the original image.

Let \textbf{\textit{L}} be our current layer, \textbf{\textit{L-1}} the previous layer and \textbf{\textit{s}} the stride. The effective window size (on the original image) for layer \textbf{\textit{L}} after downsampling, albeit using the same filter width for patch extraction can be expressed as shown in Equation~(\ref{dfg-5a33235a2cf2}) : 
\let\saveeqnno\theequation
\let\savefrac\frac
\def\dispfrac{\displaystyle\savefrac}
\begin{eqnarray}
\let\frac\dispfrac
\gdef\theequation{4}
\let\theHequation\theequation
\label{dfg-5a33235a2cf2}
\begin{array}{@{}l}Window\_size\;_L\;=\;Window\_size\;_{L-1}\;.\;s\end{array}
\end{eqnarray}
\global\let\theequation\saveeqnno
\addtocounter{equation}{-1}\ignorespaces

After the features of the input layer are correlated with the input image and downsampled, feature extraction for deeper layers is then carried the same way, repeating all the steps mentioned in this section. 

In the above designed scheme, it is assumed that the features selected in the previous section only capture defect-free information from the training image(s). It is therefore necessary to cast out any defective patches before feature selection to ensure the filters discovered do not contain abnormal elements. Training should hereby be done on defect-free images.

\bgroup
\fixFloatSize{Figures/Figure5.jpg}
\begin{figure}[!htbp]
\centering \makeatletter\IfFileExists{Figures/Figure5.jpg}{\includegraphics[width=\linewidth]{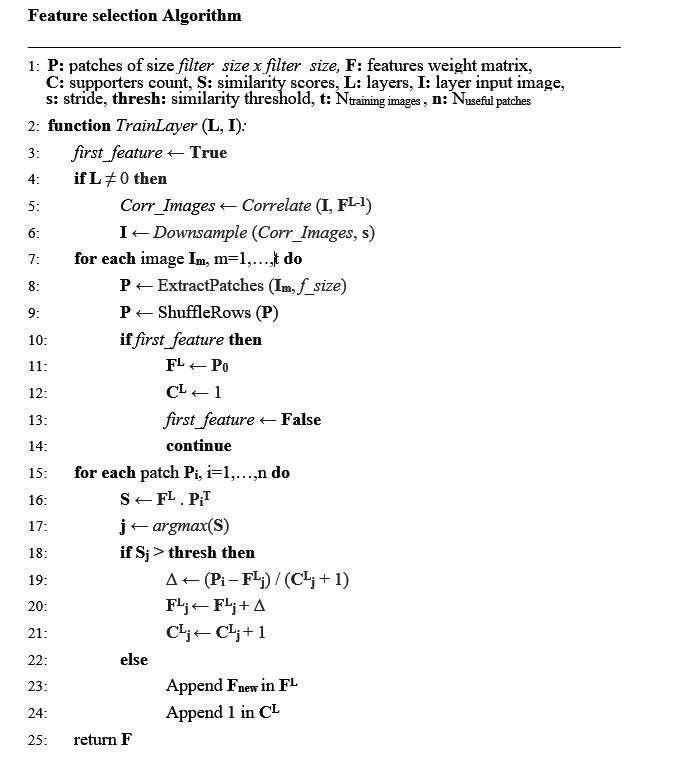}}{}
\makeatother 
\caption{{Algorithm for the feature selection step}}
\label{f-6b30a3d493fd}
\end{figure}
\egroup

\subsection{Detection of anomalies}

\subsubsection{\textit{Defect probability map }}
On the account of training on defect-free samples only, patches selected as features strictly contain defect-free information. Intuitively, the difference between these features and patches extracted from the testing image can be used as a clue for defect identification on the input image. Indeed, a distance metric, such as Manhattan distance, can be used to highlight this difference in order to detect anomalies. 

The Manhattan distance between two patches \textbf{\textit{A}} and \textbf{\textit{B}} of the same height and weight \textbf{\textit{p}} is calculated by :
\let\saveeqnno\theequation
\let\savefrac\frac
\def\dispfrac{\displaystyle\savefrac}
\begin{eqnarray}
\let\frac\dispfrac
\gdef\theequation{8}
\let\theHequation\theequation
\label{dfg-2cab24e92b77}
\begin{array}{@{}l}d(A,B)\;=\;\sum_{i=1}^{p}\sum_{j=1}^{p}\left|a_{ij}-b_{ij}\right|\end{array}
\end{eqnarray}
\global\let\theequation\saveeqnno
\addtocounter{equation}{-1}\ignorespaces 
The output of this measure is normalized to fall in the [0,1] interval. In this case, the distance score obtained is inversely similar to the similarity score used in training. It can be regarded as a dissimilarity score. 

Thereupon, we can introduce an anomaly threshold to label patches as either anomalous or anomaly-free. Here again, we focus on the highest similarity value obtained which corresponds to the lowest distance in this case. Another cut-off value intervenes at this stage to decide if a patch is defective or not. This value will be referred to as the \textit{anomaly threshold}. 

If the distance score is lower than the \textit{anomaly threshold}, it means that the patch is very similar to one of our features and hence is not an anomaly. On the contrary, if this score is higher than the \textit{anomaly threshold}, it indicates that the patch is not similar enough to any of our saved filters and that the probability of this patch being defective is high. This patch is therefore labeled as anomalous. 

Different defect probabilities are affected to the different pixels in the defective patch. Probability information is ponderously distributed in a way that it is the highest for the central pixel and the lowest in the patch's boundaries. The weight of every pixel in the patch is determined by a two-dimensional symmetric Gaussian kernel such as :
\let\saveeqnno\theequation
\let\savefrac\frac
\def\dispfrac{\displaystyle\savefrac}
\begin{eqnarray}
\let\frac\dispfrac
\gdef\theequation{9}
\let\theHequation\theequation
\label{dfg-3bb20305deaf}
\begin{array}{@{}l}G(i,\;j)\;=\;\frac1{2\pi\sigma ^{2}}\;exp\left(-\frac{i^{2}\;+\;j^{2}}{2\sigma ^{2}}\right)\end{array}
\end{eqnarray}
\global\let\theequation\saveeqnno
\addtocounter{equation}{-1}\ignorespaces 
Due to the overlapping division used in patch extraction, every single pixel is included in many patches, at different positions in every patch and every time with different neighbors. Additionally, every layer adds anomalies to construct the final anomaly output. As previously mentioned, the window size grows as we progress towards deeper layers, proportionally to the stride parameter. 

In the end, we obtain a final defect probability image is obtained where the value of every pixel indicates the certainty of that pixel's defectiveness. The higher that value, the more certain we are of that pixel being anomalous.

\subsubsection{\textit{Defect segmentation}}After we obtain an image containing defect-probability information, this image needs to be binarized in order to get a proper anomaly segmentation.  For such a binarization, finding the best threshold value is critical.

Maximum two-dimensional entropy \unskip~\cite{981091:21201211} was chosen for this step because it takes neighborhood information into consideration, as opposed to the global thresholding methods. 

Local relative entropy measures the difference of the brightness of a pixel from the mean brightness of its neighborhood:

\begin{itemize}
  \item \relax  If the brightness of a pixel is similar to its neighbours - this is usually the case when a pixel and its neighborhood belong to the same class (background class vs. anomaly class) - the local relative entropy is small.
  \item \relax On the other hand, if the brightness of a pixel is very different from that of its neighbors - this is the case when the pixel is either noise or an edge pixel - the local relative entropy is high. 
\end{itemize}
  Given a \textbf{\textit{n x n}} neighborhood with \textbf{\textit{L}} distinct grayscale intensity levels, the mean grayscale intensity level value of this neighborhood is :
\let\saveeqnno\theequation
\let\savefrac\frac
\def\dispfrac{\displaystyle\savefrac}
\begin{eqnarray}
\let\frac\dispfrac
\gdef\theequation{10}
\let\theHequation\theequation
\label{dfg-d6381740ba27}
\begin{array}{@{}l}\overline I\;(x,y)=\frac1{n^{2}}\sum_{\;\;\;i=-(n-1)/2}^{(n-1)/2}\sum_{\;\;\;j=-(n-1)/2}^{(n-1)/2}I(x+i,y+j)\end{array}
\end{eqnarray}
\global\let\theequation\saveeqnno
\addtocounter{equation}{-1}\ignorespaces 
The joint probability of pixel pairs $(i,\;j)\;=\;\left(I(x,y\right),\;\overline I(x,y)) $ is $p_{ij}\;(i,\;j\;=\;1,\;2,\;...\;,\;L-1) $.

The entropy \textbf{\textit{H}}\textit{\textbf{\ensuremath{_{B}}}} for the background region with intensity less than \textbf{\textit{(s, t) }}and the entropy \textbf{\textit{H}}\textit{\textbf{\ensuremath{_{A}}}} for the anomaly with intensity higher than \textbf{\textit{(s, t)}} are represented as :
\let\saveeqnno\theequation
\let\savefrac\frac
\def\dispfrac{\displaystyle\savefrac}
\begin{eqnarray}
\let\frac\dispfrac
\gdef\theequation{11}
\let\theHequation\theequation
\label{dfg-aef559eebf7e}
\begin{array}{@{}l}H_B(s,t)\;=\;\sum_{i=0}^{s-1}\;\sum_{j=0}^{t-1}\;\left(\frac{p_{ij}}{P_1}\right)\;\log\left(\frac{p_{ij}}{P_1}\right)\;\end{array}
\end{eqnarray}
\global\let\theequation\saveeqnno
\addtocounter{equation}{-1}\ignorespaces 
with : $P_1\;=\;\sum_{i=0}^{s-1}\;\sum_{j=0}^{t-1}\;p_{ij} $
\let\saveeqnno\theequation
\let\savefrac\frac
\def\dispfrac{\displaystyle\savefrac}
\begin{eqnarray}
\let\frac\dispfrac
\gdef\theequation{12}
\let\theHequation\theequation
\label{dfg-39bee333f21c}
\begin{array}{@{}l}H_A(s,t)\;=\;\sum_{i=s}^{L-1}\;\sum_{j=t}^{L-1}\;\left(\frac{p_{ij}}{P_2}\right)\;\log\left(\frac{p_{ij}}{P_2}\right)\end{array}
\end{eqnarray}
\global\let\theequation\saveeqnno
\addtocounter{equation}{-1}\ignorespaces 
and : $P_2\;=\;\sum_{i=s}^{L-1}\;\sum_{j=t}^{L-1}\;p_{ij} $

Then, the total entropy of the image \textbf{\textit{H}}\textit{\textbf{\ensuremath{_{I}}}} is :
\let\saveeqnno\theequation
\let\savefrac\frac
\def\dispfrac{\displaystyle\savefrac}
\begin{eqnarray}
\let\frac\dispfrac
\gdef\theequation{13}
\let\theHequation\theequation
\label{dfg-473dda1c288b}
\begin{array}{@{}l}H_{I\;}=\;H_B\;+\;H_A\end{array}
\end{eqnarray}
\global\let\theequation\saveeqnno
\addtocounter{equation}{-1}\ignorespaces 
The ideal threshold vector is \textbf{\textit{(s*, t*)}} that satisfies : 
\let\saveeqnno\theequation
\let\savefrac\frac
\def\dispfrac{\displaystyle\savefrac}
\begin{eqnarray}
\let\frac\dispfrac
\gdef\theequation{14}
\let\theHequation\theequation
\label{dfg-b0c934e9e147}
\begin{array}{@{}l}H(s\ast,\;t\ast)\;=\;max\left\{H_I(s,t)\right\}\end{array}
\end{eqnarray}
\global\let\theequation\saveeqnno
\addtocounter{equation}{-1}\ignorespaces 
Finally, opening is applied to the image to remove noise from the image, if any, while smoothing the border of the anomaly result.
    
\section{Experiments}

\subsection{Dataset}
Our experiments were performed on the Patterned Fabric dataset \unskip~\cite{981091:21201209} provided by the Industrial Automation Research Laboratory, Dept. of Electrical and Electronic Engineering of the University of Hong Kong. This dataset contains images of three kinds of patterns : dot-patterned, box-patterned and star-patterned. There are 25 defect-free samples and 25 defective samples per fabric type including broken end, hole, netting multiple, thick bar and thin bar (5 samples from each defect type). These defect types are shown in Figure~\ref{figure-fb7cbfe9a69843f88ac2097063467826}.

\subsection{Experimental setup}

\subsubsection{\textit{Training and testing sets}}We used both Box-patterned and Star-patterned sub-datasets in order to showcase our results. 

Training was performed on one defect-free sample per fabric type and testing was done on the remaining images (defect-free and defective). Training was completed in 11 seconds for the Star-patterned defect-free sample and in 10 seconds for the Box-patterned defect-free sample.

\subsubsection{\textit{Preprocessing and patch extraction}}

\begin{itemize}
  \item \relax No data augmentation technique was applied for either of the datasets;
  \item \relax Following the steps mentioned in section \textbf{3.2}. The pattern periods for the Box-patterned and Star-patterned fabric types were found to be, respectively, 26 x 24 and 21 x 16. These are shown in Figure~\ref{f-36f8fabcc024}. Considering that we are using square patches for our approach, we took the maximum of the period on the row direction and that on the column direction and added 1 to it to make it an odd number if it wasn't already. This resulted in a \textit{filter\_size} of 27 for Box-patterned and 21 for Star-patterned;
  \item \relax In both training and testing, 

\begin{itemize}
  \item \relax patches were extracted in a row by row manner as explained in section \textbf{3.3} with an overlapping of 1 pixel in both row and column directions;
  \item \relax the contrast threshold for useless patches' elimination prior to features selection was set to 0;
\end{itemize}
  
\end{itemize}
  
\bgroup
\fixFloatSize{Figures/Figure6.jpg}
\begin{figure}[!htbp]
\centering \makeatletter\IfFileExists{Figures/Figure6.jpg}{\includegraphics{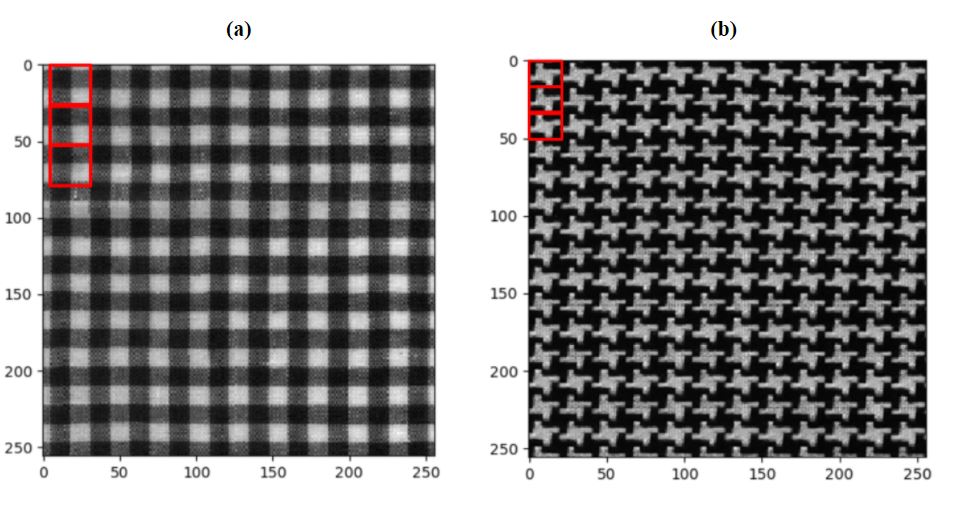}}{}
\makeatother 
\caption{{\textbf{(a)} Defect-free Box-patterned sample with 26 x 24 period ; \textbf{(b)} Defect-free Star-patterned sample with 21 x 16 period.}}
\label{f-36f8fabcc024}
\end{figure}
\egroup

\subsubsection{\textit{Anomaly threshold determination}}In order to choose the appropriate anomaly threshold and to make sure our anomaly detection will only capture defective patches, we registered the response of the network to the defect-free sample training was performed on. For every pixel of the training image, we saved the minimum distance value (the distance to the closest feature). Following that, the maximum of these values across all pixels was taken as the threshold for the anomaly detection phase described in section \textbf{3.5}.

\subsubsection{\textit{Network structure}}Experiments were conducted with a network of one layer. One epoch is enough to train the layer and no further training is needed on the filters' weights after the feature selection step explained in section \textbf{3.4.,} showcased by the algorithm in Figure~\ref{f-6b30a3d493fd}. As soon as features are selected, their weights are deemed non-trainable and are later on used in the anomaly detection step. Figure~\ref{figure-63fc439a3f18422c9e95d2a6c53b9c15} shows the details of the model that was built for experiments.
\bgroup
\fixFloatSize{Figures/Figure7.jpg}
\begin{figure*}[!htbp]
\centering \makeatletter\IfFileExists{Figures/Figure7.jpg}{\includegraphics[width=\linewidth]{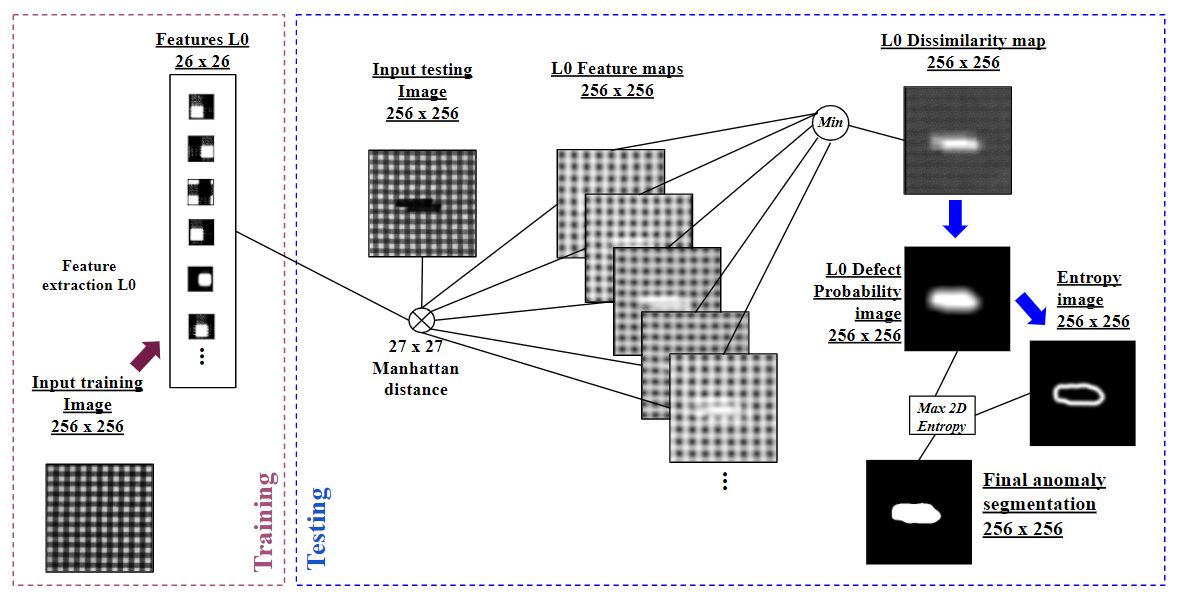}}{}
\makeatother 
\caption{{Overall Architecture for the Training and Testing phases of the Network used in the experiments. }}
\label{figure-63fc439a3f18422c9e95d2a6c53b9c15}
\end{figure*}
\egroup

\subsection{Performance measurement metrics}Ground-truth defect masks are available in the dataset. Those ground-truth images were used for the evaluation of the performance of our approach. Examples of these ground truth images are shown in Figure~\ref{figure-fb7cbfe9a69843f88ac2097063467826}. 

Each pixel in the results is then classified as either TP, TN, FP or FN by comparing it to the ground truth image with: 

\begin{itemize}
  \item \relax \textit{TP (true positives) :} number of defective pixels that are correctly identified as defective;
  \item \relax \textit{TN (true negatives) :} number of defect-free pixels that are correctly identified as defect-free;
  \item \relax \textit{FP (false positives) :} number of defect-free pixels that are incorrectly identified as defective;~
  \item \relax \textit{\textit{FN (false negatives) :}}~number of defective pixels that are incorrectly identified as defect-free.
\end{itemize}
  After that, the \textit{True positive rate (TPR), True negative rate (TNR), False negative rate (FNR), False positive rate (FPR), Positive predictive value (PPV), Accuracy (ACC) }and \textit{F1-score (F1) }evaluation metrics are computed. These metrics respectively correspond to the equations from Equation~(\ref{dfg-09bab9387149}) to Equation~(\ref{disp-formula-group-ad79c89e70a04ac5b479aa8338d65e7c}).
\let\saveeqnno\theequation
\let\savefrac\frac
\def\dispfrac{\displaystyle\savefrac}
\begin{eqnarray}
\let\frac\dispfrac
\gdef\theequation{15}
\let\theHequation\theequation
\label{dfg-09bab9387149}
\begin{array}{@{}l}TPR\;or\;Recall\;=\;\frac{TP}{TP\:+\:FN}\end{array}
\end{eqnarray}
\global\let\theequation\saveeqnno
\addtocounter{equation}{-1}\ignorespaces

\let\saveeqnno\theequation
\let\savefrac\frac
\def\dispfrac{\displaystyle\savefrac}
\begin{eqnarray}
\let\frac\dispfrac
\gdef\theequation{16}
\let\theHequation\theequation
\label{dfg-ea7038a4c7c1}
\begin{array}{@{}l}TNR\;or\;Selectivity\;=\;\frac{TN}{FP\:+\:TN}\end{array}
\end{eqnarray}
\global\let\theequation\saveeqnno
\addtocounter{equation}{-1}\ignorespaces 

\let\saveeqnno\theequation
\let\savefrac\frac
\def\dispfrac{\displaystyle\savefrac}
\begin{eqnarray}
\let\frac\dispfrac
\gdef\theequation{17}
\let\theHequation\theequation
\label{dfg-8cc8afe046c8}
\begin{array}{@{}l}FNR\;or\;Missed\;detection\;rate\;=\;\frac{FN}{TP\:+\:FN}\end{array}
\end{eqnarray}
\global\let\theequation\saveeqnno
\addtocounter{equation}{-1}\ignorespaces 

\let\saveeqnno\theequation
\let\savefrac\frac
\def\dispfrac{\displaystyle\savefrac}
\begin{eqnarray}
\let\frac\dispfrac
\gdef\theequation{18}
\let\theHequation\theequation
\label{dfg-fd2e448a99e0}
\begin{array}{@{}l}FPR\;or\;False\;alarm\;rate=\frac{FP}{FP\:+\:TN}\;\end{array}
\end{eqnarray}
\global\let\theequation\saveeqnno
\addtocounter{equation}{-1}\ignorespaces 

\let\saveeqnno\theequation
\let\savefrac\frac
\def\dispfrac{\displaystyle\savefrac}
\begin{eqnarray}
\let\frac\dispfrac
\gdef\theequation{19}
\let\theHequation\theequation
\label{dfg-869ba968230c}
\begin{array}{@{}l}PPV\;or\;Precision\;=\frac{TP}{TP\:+\:FP}\;\end{array}
\end{eqnarray}
\global\let\theequation\saveeqnno
\addtocounter{equation}{-1}\ignorespaces 

\let\saveeqnno\theequation
\let\savefrac\frac
\def\dispfrac{\displaystyle\savefrac}
\begin{eqnarray}
\let\frac\dispfrac
\gdef\theequation{20}
\let\theHequation\theequation
\label{dfg-1ccf68c8052b}
\begin{array}{@{}l}ACC\;or\;DSR\;=\frac{TP\:+\:TN}{TP\:+\:TN\:+\:FP\:+\:FN}\;\end{array}
\end{eqnarray}
\global\let\theequation\saveeqnno
\addtocounter{equation}{-1}\ignorespaces 

\let\saveeqnno\theequation
\let\savefrac\frac
\def\dispfrac{\displaystyle\savefrac}
\begin{eqnarray}
\let\frac\dispfrac
\gdef\theequation{21}
\let\theHequation\theequation
\label{disp-formula-group-ad79c89e70a04ac5b479aa8338d65e7c}
\begin{array}{@{}l}F1=2\;.\;\frac{Precision\;.\;Recall}{Precision+Recall}\end{array}
\end{eqnarray}
\global\let\theequation\saveeqnno
\addtocounter{equation}{-1}\ignorespaces 

\bgroup
\fixFloatSize{Figures/Figure8.png}
\begin{figure*}[!htbp]
\centering \makeatletter\IfFileExists{Figures/Figure8.png}{\includegraphics[width=.72\linewidth]{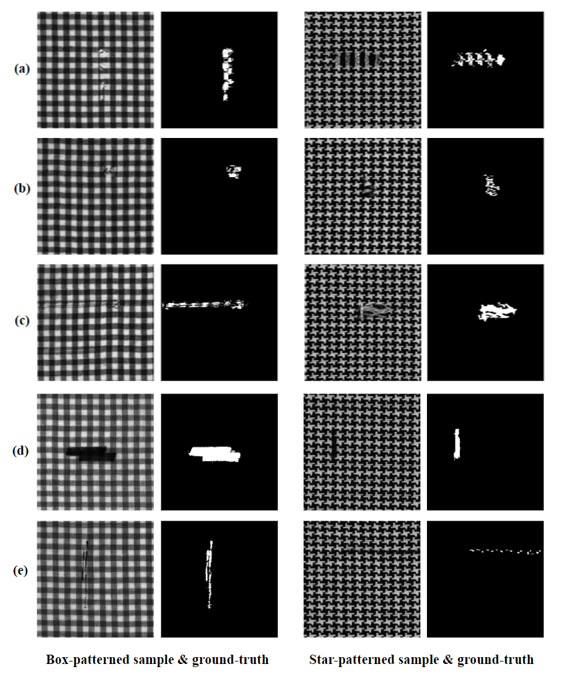}}{}
\makeatother 
\caption{{Illustration of Box-patterned and Star-patterned samples with their ground-truth images. \textbf{(a)} Broken End; \textbf{(b)} Hole; \textbf{(c)} Netting multiple; \textbf{(d)} Thick Bar and \textbf{(e)} Thin Bar.}}
\label{figure-fb7cbfe9a69843f88ac2097063467826}
\end{figure*}
\egroup
    
\section{Results and discussion}
Illustrations of the anomaly detection output can be seen in Figure~\ref{f-64a6ad7d2995}. The corresponding scores of the experiments run with the above mentioned parameters will be explained in detail in section \textbf{5.2.}
\bgroup
\fixFloatSize{Figures/Figure9.png}
\begin{figure*}[!htbp]
\centering 
\makeatletter\IfFileExists{Figures/Figure9.png}{\includegraphics[width=.72\linewidth]{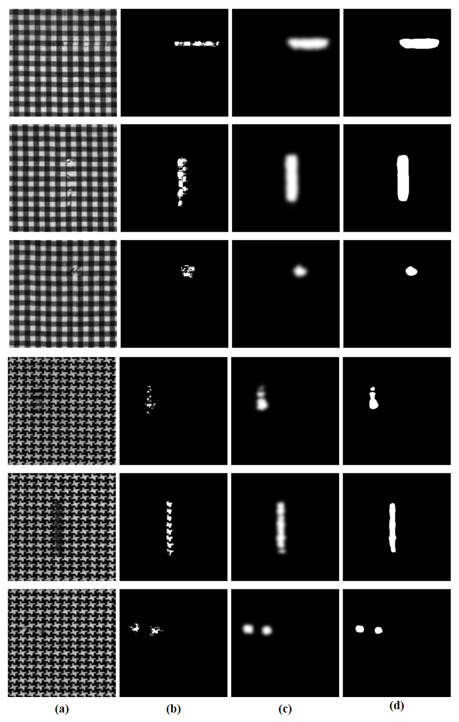}}{}
\makeatother 
\caption{{Examples of anomaly detection results on the Box-patterned and Star-patterned samples : \textbf{(a) }input image; \textbf{(b)} ground truth image; \textbf{(c)} weighted defect probability map after anomaly detection; \textbf{(d)} final binarized result after 2D Max Entropy. }}
\label{f-64a6ad7d2995}
\end{figure*}
\egroup

\subsection{Features selected}The number of features extracted essentially depends on the fabric type. The average number of filters selected in training with threshold =0.7 was 320 on the Box-patterned reference image and 451 on the Star-patterned reference image. 

Examples of obtained features for both Fabric types can be seen in Figure~\ref{f-aeb5faa4a5e2}. We can see that all learned filters correspond to intuitive representations of the input domain and exhaustively represent the fabric textures in their different variations.

\bgroup
\fixFloatSize{Figures/Figure10.jpg}
\begin{figure}[!htbp]
\centering \makeatletter\IfFileExists{Figures/Figure10.jpg}{\includegraphics[width=\linewidth]{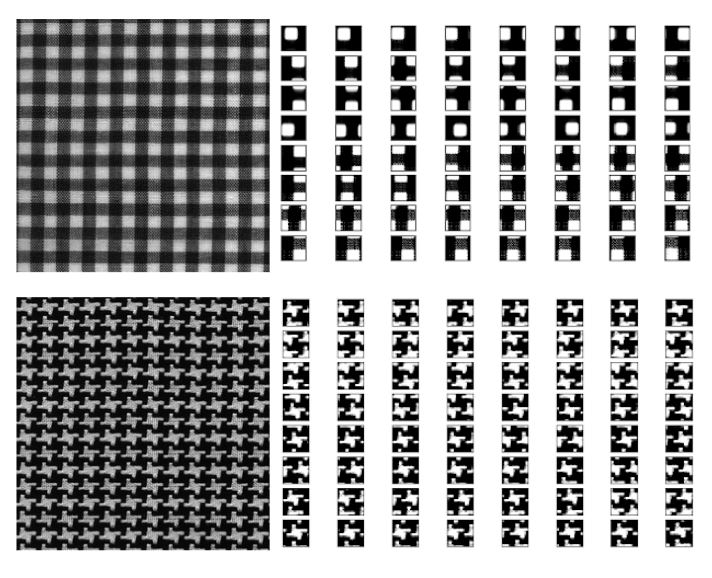}}{}
\makeatother 
\caption{{Illustration of features extracted from the two training samples used, Box-patterned and Star-patterned.}}
\label{f-aeb5faa4a5e2}
\end{figure}
\egroup

\subsection{Performance evaluation}
Our model's performance by defect type and by fabric type is summarized, for the Box-patterned samples and the Star-patterned samples, by the scores respectively displayed in Table~\ref{tw-091ea4645fe6} and Table~\ref{tw-9df04ce4be09}. We showcase the performance of our proposed approach compared to the other approaches that used the unsupervised learning paradigm. The methods were compared on the same benchmark Star-patterned fabric samples and are shown in tables from Table~\ref{tw-9df04ce4be09} to Table~\ref{tw-6e806ca05334}, where the best scores are highlighted in bold. The approaches we compare to are (in chronological order) Wavelet Golden Image Subtraction (WGIS)\unskip~\cite{981091:21201168}, Bollinger Bands (BB)\unskip~\cite{981091:21201212}, Regular Bands (RB)\unskip~\cite{981091:21201213}, Image Decomposition (ID)\unskip~\cite{981091:21201214}, Elo Rating method (ER)\unskip~\cite{981091:21201215}, Stacked Denoising Auto-Encoder (SDAE)\unskip~\cite{981091:21201196}, Fisher Criterion-Based Stacked Denoising Auto-Encoder (FCSDA)\unskip~\cite{981091:21201196}, Multi-Scale Convolutional Denoising Auto-Encoder (MSCDAE)\unskip~\cite{981091:21201198}  and Cross-Patch Similarity Auto-encoder (CPSAE)\unskip~\cite{981091:21201169} .

\begin{table}[!htbp]
\caption{{\textbf{\textit{Overall}} scores on the \textit{\textbf{Box-patterned}} fabric type per defect type} }
\label{tw-091ea4645fe6}
\def\arraystretch{1}
\ignorespaces 
\centering 
\begin{tabulary}{\linewidth}{p{\dimexpr.2848\linewidth-2\tabcolsep}p{\dimexpr.1152\linewidth-2\tabcolsep}p{\dimexpr.16009999999999998\linewidth-2\tabcolsep}p{\dimexpr.22330000000000002\linewidth-2\tabcolsep}p{\dimexpr.2166\linewidth-2\tabcolsep}}
\tbltoprule \rAlignHack Defect type & \cAlignHack DSR & \cAlignHack Recall & \cAlignHack Precision & \cAlignHack F1-score\\
\tblmidrule 
\rAlignHack Broken End &
  \cAlignHack 0.98 &
  \cAlignHack 0.81 &
  \cAlignHack 0.54 &
  \cAlignHack 0.65\\
\rAlignHack Hole &
  \cAlignHack 0.99 &
  \cAlignHack 0.37 &
  \cAlignHack 0.63 &
  \cAlignHack 0.47\\
\rAlignHack Netting Multiple &
  \cAlignHack 0.98 &
  \cAlignHack 0.41 &
  \cAlignHack 0.42 &
  \cAlignHack 0.42\\
\rAlignHack Thick Bar &
  \cAlignHack 0.98 &
  \cAlignHack 0.90 &
  \cAlignHack 0.67 &
  \cAlignHack 0.77\\
\rAlignHack Thin Bar &
  \cAlignHack 0.97 &
  \cAlignHack 0.58 &
  \cAlignHack 0.25 &
  \cAlignHack 0.35\\
\rAlignHack Defect-free &
  \cAlignHack 1.00 &
  \cAlignHack N/A &
  \cAlignHack N/A &
  \cAlignHack N/A\\
\rAlignHack Overall &
  \cAlignHack 0.98 &
  \cAlignHack 0.70 &
  \cAlignHack 0.52 &
  \cAlignHack 0.60\\
\tblbottomrule 
\end{tabulary}\par 
\end{table}

\begin{table}[!htbp]
\caption{{\textbf{\textit{Overall}} scores on the \textit{\textbf{Star-patterned}} fabric type} }
\label{tw-9df04ce4be09}
\def\arraystretch{1}
\ignorespaces 
\centering 
\begin{tabulary}{\linewidth}{p{\dimexpr.21500000000000004\linewidth-2\tabcolsep}p{\dimexpr.18499999999999996\linewidth-2\tabcolsep}p{\dimexpr.20\linewidth-2\tabcolsep}p{\dimexpr.20\linewidth-2\tabcolsep}p{\dimexpr.20\linewidth-2\tabcolsep}}
\tbltoprule \rAlignHack Method & \cAlignHack DSR & \cAlignHack Recall & \cAlignHack Precision & \cAlignHack F1-score\\
\tblmidrule 
\rAlignHack WGIS  &
  \cAlignHack 0.96 &
  \cAlignHack 0.01 &
  \cAlignHack 0.01 &
  \cAlignHack 0.01\\
\rAlignHack BB   &
  \cAlignHack 0.98 &
  \cAlignHack 0.11 &
  \cAlignHack 0.27 &
  \cAlignHack 0.15\\
\rAlignHack RB  &
  \cAlignHack 0.98 &
  \cAlignHack 0.10 &
  \cAlignHack 0.37 &
  \cAlignHack 0.15\\
\rAlignHack ID  &
  \cAlignHack 0.98 &
  \cAlignHack 0.81 &
  \cAlignHack 0.40 &
  \cAlignHack 0.50\\
\rAlignHack ER &
  \cAlignHack \textbf{0.99} &
  \cAlignHack 0.33 &
  \cAlignHack 0.20 &
  \cAlignHack 0.25\\
\rAlignHack SDAE  &
  \cAlignHack 0.98 &
  \cAlignHack 0.38 &
  \cAlignHack 0.32 &
  \cAlignHack 0.31\\
\rAlignHack FCSDA &
  \cAlignHack 0.99 &
  \cAlignHack 0.64 &
  \cAlignHack 0.47 &
  \cAlignHack 0.52\\
\rAlignHack MSCDAE &
  \cAlignHack 0.98 &
  \cAlignHack 0.30 &
  \cAlignHack 0.32 &
  \cAlignHack 0.30\\
\rAlignHack CPSAE &
  \cAlignHack \textbf{0.99} &
  \cAlignHack 0.66 &
  \cAlignHack 0.50 &
  \cAlignHack 0.55\\
\rAlignHack Ours &
  \cAlignHack \textbf{0.99} &
  \cAlignHack \textbf{0.85} &
  \cAlignHack \textbf{0.71} &
  \cAlignHack \textbf{0.77}\\
\tblbottomrule 
\end{tabulary}\par 
\end{table}

\begin{table}[!htbp]
\caption{{Scores for the \textbf{\textit{Broken End}} defect type on \textbf{\textit{Star-patterned}} fabric type} }
\label{tw-a6f81addd5db}
\def\arraystretch{1}
\ignorespaces 
\centering 
\begin{tabulary}{\linewidth}{p{\dimexpr.33959999999999994\linewidth-2\tabcolsep}p{\dimexpr.1401\linewidth-2\tabcolsep}p{\dimexpr.14189999999999996\linewidth-2\tabcolsep}p{\dimexpr.1767\linewidth-2\tabcolsep}p{\dimexpr.20169999999999998\linewidth-2\tabcolsep}}
\tbltoprule \rAlignHack Method & \cAlignHack DSR & \cAlignHack Recall & \cAlignHack Precision & \cAlignHack F1-score\\
\tblmidrule 
\rAlignHack WGIS &
  \cAlignHack 0.96 &
  \cAlignHack 0.00 &
  \cAlignHack 0.00 &
  \cAlignHack 0.00\\
\rAlignHack BB &
  \cAlignHack 0.98 &
  \cAlignHack 0.03 &
  \cAlignHack 0.03 &
  \cAlignHack 0.03\\
\rAlignHack RB &
  \cAlignHack \textbf{0.99} &
  \cAlignHack 0.05 &
  \cAlignHack 0.08 &
  \cAlignHack 0.06\\
\rAlignHack ID &
  \cAlignHack 0.98 &
  \cAlignHack 0.76 &
  \cAlignHack 0.18 &
  \cAlignHack 0.29\\
\rAlignHack ER &
  \cAlignHack 0.98 &
  \cAlignHack 0.09 &
  \cAlignHack 0.07 &
  \cAlignHack 0.08\\
\rAlignHack SDAE &
  \cAlignHack \textbf{0.99} &
  \cAlignHack 0.74 &
  \cAlignHack 0.24 &
  \cAlignHack 0.36\\
\rAlignHack FCSDA &
  \cAlignHack \textbf{0.99} &
  \cAlignHack 0.70 &
  \cAlignHack 0.25 &
  \cAlignHack 0.37\\
\rAlignHack MSCDAE &
  \cAlignHack \textbf{0.99} &
  \cAlignHack 0.35 &
  \cAlignHack 0.25 &
  \cAlignHack 0.29\\
\rAlignHack CPSAE &
  \cAlignHack \textbf{0.99} &
  \cAlignHack 0.77 &
  \cAlignHack 0.40 &
  \cAlignHack 0.53\\
\rAlignHack Ours &
  \cAlignHack \textbf{0.99} &
  \cAlignHack \textbf{0.85} &
  \cAlignHack \textbf{0.40} &
  \cAlignHack \textbf{0.54}\\
\tblbottomrule 
\end{tabulary}\par 
\end{table}

\begin{table}[!htbp]
\caption{{Scores for the \textbf{\textit{Hole}} defect type on \textbf{\textit{Star-patterned}} fabric type} }
\label{tw-cdf7fb312b96}
\def\arraystretch{1}
\ignorespaces 
\centering 
\begin{tabulary}{\linewidth}{LLLLL}
\tbltoprule \rAlignHack Method & \cAlignHack DSR & \cAlignHack Recall & \cAlignHack Precision & \cAlignHack F1-score\\
\tblmidrule 
\rAlignHack WGIS &
  \cAlignHack 0.96 &
  \cAlignHack 0.00 &
  \cAlignHack 0.00 &
  \cAlignHack 0.00\\
\rAlignHack BB &
  \cAlignHack \textbf{0.99} &
  \cAlignHack 0.03 &
  \cAlignHack 0.08 &
  \cAlignHack 0.04\\
\rAlignHack RB &
  \cAlignHack \textbf{0.99} &
  \cAlignHack 0.07 &
  \cAlignHack 0.26 &
  \cAlignHack 0.11\\
\rAlignHack ID &
  \cAlignHack 0.97 &
  \cAlignHack 0.72 &
  \cAlignHack 0.33 &
  \cAlignHack 0.46\\
\rAlignHack ER &
  \cAlignHack 0.98 &
  \cAlignHack 0.24 &
  \cAlignHack 0.12 &
  \cAlignHack 0.16\\
\rAlignHack SDAE &
  \cAlignHack \textbf{0.99} &
  \cAlignHack 0.31 &
  \cAlignHack 0.20 &
  \cAlignHack 0.24\\
\rAlignHack FCSDA &
  \cAlignHack \textbf{0.99} &
  \cAlignHack 0.11 &
  \cAlignHack 0.14 &
  \cAlignHack 0.12\\
\rAlignHack MSCDAE &
  \cAlignHack \textbf{0.99} &
  \cAlignHack 0.36 &
  \cAlignHack 0.28 &
  \cAlignHack 0.32\\
\rAlignHack CPSAE &
  \cAlignHack \textbf{0.99} &
  \cAlignHack 0.71 &
  \cAlignHack 0.40 &
  \cAlignHack 0.51\\
\rAlignHack Ours &
  \cAlignHack \textbf{0.99} &
  \cAlignHack \textbf{0.81} &
  \cAlignHack \textbf{0.49} &
  \cAlignHack \textbf{0.62}\\
\tblbottomrule 
\end{tabulary}\par 
\end{table}

\begin{table}[!htbp]
\caption{{Scores for the \textbf{\textit{Netting Multiple}} defect type on \textbf{\textit{Star-patterned}} fabric type} }
\label{tw-46379712333b}
\def\arraystretch{1}
\ignorespaces 
\centering 
\begin{tabulary}{\linewidth}{LLLLL}
\tbltoprule \rAlignHack Method & \cAlignHack DSR & \cAlignHack Recall & \cAlignHack Precision & \cAlignHack F1-score\\
\tblmidrule 
\rAlignHack WGIS &
  \cAlignHack 0.95 &
  \cAlignHack 0.01 &
  \cAlignHack 0.01 &
  \cAlignHack 0.01\\
\rAlignHack BB &
  \cAlignHack 0.98 &
  \cAlignHack 0.12 &
  \cAlignHack 0.37 &
  \cAlignHack 0.18\\
\rAlignHack RB &
  \cAlignHack 0.98 &
  \cAlignHack 0.15 &
  \cAlignHack 0.48 &
  \cAlignHack 0.23\\
\rAlignHack ID &
  \cAlignHack \textbf{0.99} &
  \cAlignHack 0.82 &
  \cAlignHack 0.57 &
  \cAlignHack 0.67\\
\rAlignHack ER &
  \cAlignHack 0.98 &
  \cAlignHack 0.16 &
  \cAlignHack 0.13 &
  \cAlignHack 0.14\\
\rAlignHack SDAE &
  \cAlignHack 0.98 &
  \cAlignHack 0.08 &
  \cAlignHack 0.21 &
  \cAlignHack 0.12\\
\rAlignHack FCSDA &
  \cAlignHack \textbf{0.99} &
  \cAlignHack 0.14 &
  \cAlignHack 0.32 &
  \cAlignHack 0.20\\
\rAlignHack MSCDAE &
  \cAlignHack 0.98 &
  \cAlignHack 0.34 &
  \cAlignHack 0.45 &
  \cAlignHack 0.39\\
\rAlignHack CPSAE &
  \cAlignHack \textbf{0.99} &
  \cAlignHack 0.61 &
  \cAlignHack 0.62 &
  \cAlignHack 0.62\\
\rAlignHack Ours &
  \cAlignHack \textbf{0.99} &
  \cAlignHack \textbf{0.90} &
  \cAlignHack \textbf{0.64} &
  \cAlignHack \textbf{0.75}\\
\tblbottomrule 
\end{tabulary}\par 
\end{table}

\begin{table}[!htbp]
\caption{{Scores for the \textbf{\textit{Thick Bar}} defect type on \textbf{\textit{Star-patterned}} fabric type} }
\label{tw-1861c42780f7}
\def\arraystretch{1}
\ignorespaces 
\centering 
\begin{tabulary}{\linewidth}{LLLLL}
\tbltoprule \rAlignHack Method & \cAlignHack DSR & \cAlignHack Recall & \cAlignHack Precision & \cAlignHack F1-score\\
\tblmidrule 
\rAlignHack WGIS &
  \cAlignHack 0.93 &
  \cAlignHack 0.04 &
  \cAlignHack 0.05 &
  \cAlignHack 0.05\\
\rAlignHack BB &
  \cAlignHack 0.97 &
  \cAlignHack 0.31 &
  \cAlignHack 0.63 &
  \cAlignHack 0.41\\
\rAlignHack RB &
  \cAlignHack 0.97 &
  \cAlignHack 0.20 &
  \cAlignHack 0.77 &
  \cAlignHack 0.31\\
\rAlignHack ID &
  \cAlignHack \textbf{0.99} &
  \cAlignHack \textbf{0.94} &
  \cAlignHack 0.80 &
  \cAlignHack \textbf{0.87}\\
\rAlignHack ER &
  \cAlignHack 0.97 &
  \cAlignHack 0.70 &
  \cAlignHack 0.55 &
  \cAlignHack 0.61\\
\rAlignHack SDAE &
  \cAlignHack \textbf{0.99} &
  \cAlignHack 0.70 &
  \cAlignHack 0.94 &
  \cAlignHack 0.80\\
\rAlignHack FCSDA &
  \cAlignHack \textbf{0.99} &
  \cAlignHack 0.63 &
  \cAlignHack \textbf{0.98} &
  \cAlignHack 0.77\\
\rAlignHack MSCDAE &
  \cAlignHack 0.96 &
  \cAlignHack 0.16 &
  \cAlignHack 0.36 &
  \cAlignHack 0.22\\
\rAlignHack CPSAE &
  \cAlignHack 0.97 &
  \cAlignHack 0.45 &
  \cAlignHack 0.56 &
  \cAlignHack 0.50\\
\rAlignHack Ours &
  \cAlignHack \textbf{0.99} &
  \cAlignHack 0.81 &
  \cAlignHack 0.91 &
  \cAlignHack 0.86\\
\tblbottomrule 
\end{tabulary}\par 
\end{table}

\begin{table}[!htbp]
\caption{{Scores for the \textbf{\textit{Thin Bar}} defect type on \textbf{\textit{Star-patterned}} fabric type} }
\label{tw-6e806ca05334}
\def\arraystretch{1}
\ignorespaces 
\centering 
\begin{tabulary}{\linewidth}{LLLLL}
\tbltoprule \rAlignHack Method & DSR & Recall & Precision & F1-score\\
\tblmidrule 
\rAlignHack WGIS &
  \cAlignHack 0.95 &
  \cAlignHack 0.00 &
  \cAlignHack 0.00 &
  \cAlignHack 0.00\\
\rAlignHack BB &
  \cAlignHack \textbf{0.99} &
  \cAlignHack 0.07 &
  \cAlignHack 0.22 &
  \cAlignHack 0.10\\
\rAlignHack RB &
  \cAlignHack \textbf{0.99} &
  \cAlignHack 0.03 &
  \cAlignHack 0.28 &
  \cAlignHack 0.06\\
\rAlignHack ID &
  \cAlignHack 0.95 &
  \cAlignHack \textbf{0.82} &
  \cAlignHack 0.14 &
  \cAlignHack 0.24\\
\rAlignHack ER &
  \cAlignHack 0.97 &
  \cAlignHack 0.45 &
  \cAlignHack 0.13 &
  \cAlignHack 0.20\\
\rAlignHack SDAE &
  \cAlignHack 0.98 &
  \cAlignHack 0.31 &
  \cAlignHack 0.22 &
  \cAlignHack 0.25\\
\rAlignHack FCSDA &
  \cAlignHack 0.98 &
  \cAlignHack 0.57 &
  \cAlignHack 0.34 &
  \cAlignHack 0.43\\
\rAlignHack MSCDAE &
  \cAlignHack 0.98 &
  \cAlignHack 0.32 &
  \cAlignHack 0.26 &
  \cAlignHack 0.29\\
\rAlignHack CPSAE &
  \cAlignHack \textbf{0.99} &
  \cAlignHack 0.74 &
  \cAlignHack 0.50 &
  \cAlignHack 0.60\\
\rAlignHack Ours &
  \cAlignHack \textbf{0.99} &
  \cAlignHack 0.81 &
  \cAlignHack \textbf{0.91} &
  \cAlignHack \textbf{0.86}\\
\tblbottomrule 
\end{tabulary}\par 
\end{table}
Our method achieves the best overall detection performance as presented in Table~\ref{tw-9df04ce4be09} across all other methods. In Table~\ref{tw-a6f81addd5db} to Table~\ref{tw-6e806ca05334}, it can be seen that our approach achieves the highest scores and hence performs far better than the other approaches on most defect types with a high accuracy rate and higher recall, precision and f1-score. Moreover, all defect-free images were detected correctly for both fabric types.

On the Thick Bar and Thin Bar defect types, only the ID method has a better recall rate but at the cost of a lower precision (much lower on the Thin Bar defect type). With a rather close recall rate, and a much higher precision, our method achieves a good compromise between precision and recall which is showcased by a comparable f1 score on the Thick Bar defect type and a considerably higher f1-score on the Thin Bar defect type.

To visually distinguish the effectiveness of our approach, we plot the ROC curve and the Precision-Recall curves on both Box-patterned and Star-patterned fabric types in Figure~\ref{f-4fd8a7ce772c} and Figure~\ref{f-b334244f0f6b}. It can be noticed that the chosen value in the anomaly threshold determination in section \textbf{4.2.d.} , namely 0.5, is the one that gives the best trade-off between Recall and Precision and between Recall and FPR.

\bgroup
\fixFloatSize{Figures/Figure11.jpg}
\begin{figure*}[!htbp]
\centering \makeatletter\IfFileExists{Figures/Figure11.jpg}{\includegraphics{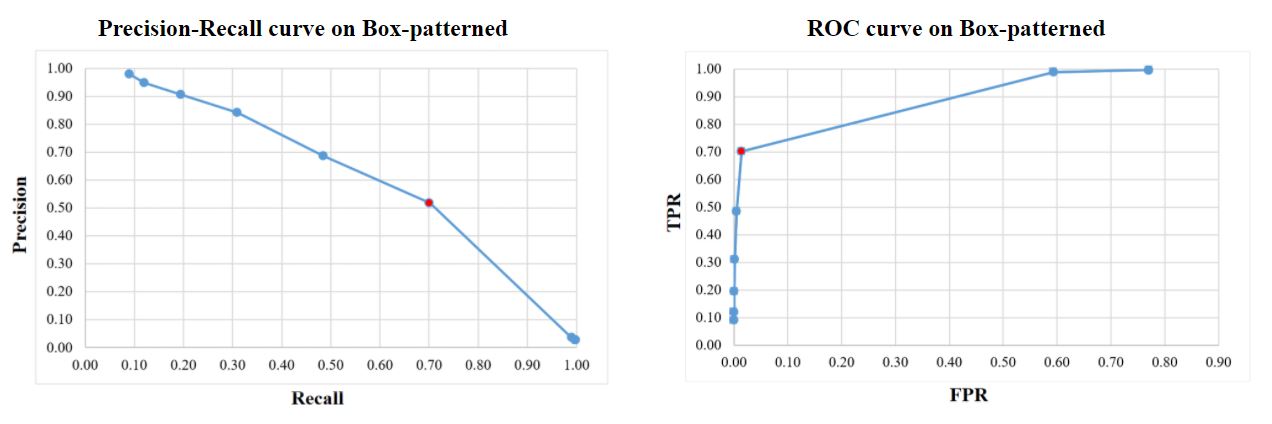}}{}
\makeatother 
\caption{{Precision-Recall and ROC curve on the Box-patterned fabric type with 0.7 as a training threshold and different thresholds for anomaly detection (with a step of 0.1).}}
\label{f-4fd8a7ce772c}
\end{figure*}
\egroup

\bgroup
\fixFloatSize{Figures/Figure12.jpg}
\begin{figure*}[!htbp]
\centering \makeatletter\IfFileExists{Figures/Figure12.jpg}{\includegraphics{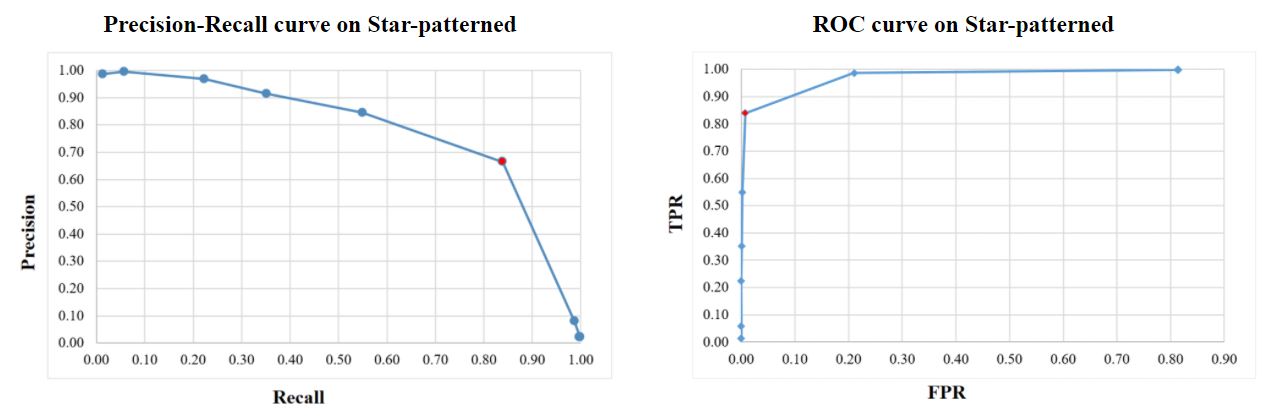}}{}
\makeatother 
\caption{{Precision-Recall and ROC curve on the Star-patterned fabric type with 0.7 as a training threshold and different thresholds for anomaly detection (with a step of 0.1).}}
\label{f-b334244f0f6b}
\end{figure*}
\egroup

\subsection{The effect of parameters}When the similarity threshold is too small (below 0.5), fewer patches are selected as features in training. This results in an under fitting on the training data. Consequently, false negatives increase in the anomaly detection phase. When this threshold is too large (usually 0.75 or higher), it leads to overfitting the training data. This increases the number of false positives in anomaly detection. 

When the anomaly threshold is too high compared to the threshold, fewer patches of the input testing image are labeled as defective. This leads to an increase of false negatives. On the other hand, if this threshold is much lower than the similarity threshold, this causes false positives to increase since more patches are deemed defective.

\bgroup
\fixFloatSize{Figures/Figure13.jpg}
\begin{figure*}[!htbp]
\centering \makeatletter\IfFileExists{Figures/Figure13.jpg}{\includegraphics{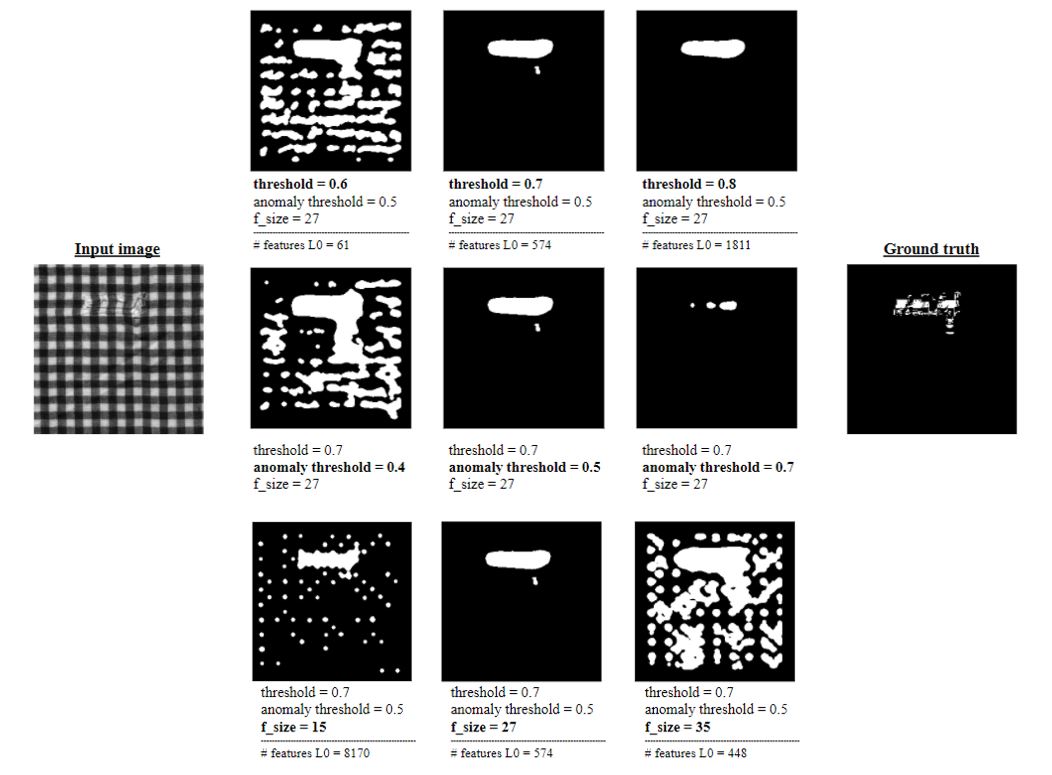}}{}
\makeatother 
\caption{{Illustration of the impact of the parameters: threshold, anomaly threshold and filter size. }}
\label{f-20970d0a9e86}
\end{figure*}
\egroup
When the filter size is smaller than the pattern period, the selected features fail to capture sufficient discriminative information for representing fabric textures. Moreover, when this filter size is smaller than the defect size in particular, the model's ability to distinguish defects declines. On the contrary, when the filter size is larger than the pattern period, selected features end up containing more complex texture information and this also affects the model's performance negatively. The impact of all these parameters is illustrated in Figure~\ref{f-20970d0a9e86}.

\subsection{Analysis of the computational cost}To further showcase the efficiency of our algorithm, it is compared against the other deep-learning based methods that have been used for fabric defect detection. 

Table~\ref{tw-60ff1fbebddb} summarizes the required time for training of the DCGAN, MSCDAE and SDAE, taken from\unskip~\cite{981091:21201204} and \unskip~\cite{981091:21201196} compared against ours. Our method is the least costly and only takes 11 seconds for training.

\begin{table*}[!htbp]
\caption{{Time performance comparison} }
\label{tw-60ff1fbebddb}
\def\arraystretch{1}
\ignorespaces 
\centering 
\begin{tabulary}{\linewidth}{p{\dimexpr.220999999999\linewidth-2\tabcolsep}p{\dimexpr.1059\linewidth-2\tabcolsep}p{\dimexpr.11840000000000002\linewidth-2\tabcolsep}p{\dimexpr.128199999999\linewidth-2\tabcolsep}p{\dimexpr.239\linewidth-2\tabcolsep}p{\dimexpr.1875\linewidth-2\tabcolsep}}
\tbltoprule  & \cAlignHack DCGAN & \cAlignHack MSCDAE & \cAlignHack SDAE & \cAlignHack FCSDAE & \cAlignHack Ours\\
\tblmidrule 
\rAlignHack No. of layers \mbox{}\protect\newline  &
  \cAlignHack 15 &
  \cAlignHack 9 &
  \cAlignHack 4 &
  \cAlignHack 8 &
  \cAlignHack 1\\
\rAlignHack Specifications of computer used  \mbox{}\protect\newline  &
  \multicolumn{3}{p{\dimexpr(.3525\linewidth-2\tabcolsep)}}{\cAlignHack Xeon E5-2640 processor and 32GB memory, NVIDIA Tesla P4 GPU using CUDA 8.0, Python 3.6 with of TensorFlow} &
  \cAlignHack Intel i5 processor 3.2 GHz and 8.0 GB memory, MATLAB R2012B &
  \cAlignHack Intel i7-6820HQ 2.70GHz and 32GB memory \\
Training time (s) &
  \cAlignHack 840.6 &
  \cAlignHack 169.8 &
  \cAlignHack 162 &
  \cAlignHack 296.9 &
  \cAlignHack 11\\
\tblbottomrule 
\end{tabulary}\par 
\end{table*}
Additionally, the consumption of computing resources and training time can be quantified by the number of parameters (trainable parameters). Table~\ref{tw-3ec14cb9d1af} shows a comparison of model efficiency of our method against the state-of-the-art CNN-based methods that have been used for fabric anomaly detection, namely FCN\unskip~\cite{981091:21201181}, SegNet\unskip~\cite{981091:21201173}, U-Net\unskip~\cite{981091:21201172}, PTIP\unskip~\cite{981091:21201221} and Mobile U-Net\unskip~\cite{981091:21201220}. As shown in Table~\ref{tw-3ec14cb9d1af}, our model only uses only one layer to achieve the results reported and has approximately 1/10000 fewer parameters than the method that has the least number of parameters while the other methods have more than four times as many parameters. As explained in section \textbf{3.4.} the number of filters in our approach is not a user-set parameter, but is rather dynamic and is found with respect to the input domain (here, the fabric type). The number of features dynamically found for Box-patterned and Star-patterned was, respectively, 320 and 451, as reported in section \textbf{5.1}. The number of parameters reported for our method in Table~\ref{tw-3ec14cb9d1af} is the highest, which is that of training on the Star-patterned fabric type.

\begin{table*}[!htbp]
\caption{{Comparison of model efficiency} }
\label{tw-3ec14cb9d1af}
\def\arraystretch{1}
\ignorespaces 
\centering 
\begin{tabulary}{\linewidth}{LLLLLLL}
\tbltoprule  & \cAlignHack FCN & \cAlignHack SegNet & \cAlignHack U-Net & \cAlignHack PTIP & \cAlignHack Mobile-UNet & \cAlignHack Ours\\
\tblmidrule 
\rAlignHack No. of layers &
  \cAlignHack 38 &
  \cAlignHack 36 &
  \cAlignHack 26 &
  \cAlignHack 10 &
  \cAlignHack 9 &
  \cAlignHack 1\\
\rAlignHack No. of parameters &
  \cAlignHack 18.6M &
  \cAlignHack 29.4M &
  \cAlignHack 31.1M &
  \cAlignHack 7.15M &
  \cAlignHack 4.6M &
  \cAlignHack 451\\
\tblbottomrule 
\end{tabulary}\par 
\end{table*}

\section{Conclusions}
We proposed a novel motif-based approach for fabric defect detection. Thanks to its dynamic and input domain-based filter selection, our method is the first unsupervised method to capitalize on the strengths of convolutional neural networks without the limitations posed by the backpropagation mechanism while saving time and effort on the initialiazition of hyperparameters and their fine-tuning. Our approach doesn't require labeled data or defective samples for training. By training on only one reference image per fabric type and converging in only one epoch, it achieves far better results than the state-of-the-art unsupervised methods on all defect types on recall, precision and f1-measure. Furthermore, it allows efficient training with a lower computational cost and in less time, compared to the most efficient CNN-based approaches.

The proposed approach requires a defect-free reference image per fabric type for training.  In the future, we intend to perform automatic elimination of defective patches based on the number of supporters that indicates the frequency of observations, rather than using only defect free images for training. 
    
\section{Acknowledgments}
This research is part of project ``Competitive Deep Learning with Convolutional Neural Networks", grant number 118E293, supported by The Support Programme for Scientific and Technological Research Projects (1001) of The Scientific and Technological Research Council of Turkey (T{\"{U}}B{\.{I}}TAK).

\bibliographystyle{elsarticle-num-names} 
\bibliography{references}

\end{document}